\documentclass{IEEEtran}

\usepackage{lineno}
\usepackage{epsfig}
\usepackage{threeparttable}
\usepackage{array}
\usepackage[usenames, dvipsnames]{color}

\ifCLASSINFOpdf

\usepackage{amsmath}
\usepackage{mathtools}
\usepackage{bm}
\DeclarePairedDelimiter\ceil{\lceil}{\rceil}
\DeclarePairedDelimiter\floor{\lfloor}{\rfloor}

\DeclareMathOperator*{\argmax}{\arg\!\max}

\DeclarePairedDelimiter\norm{\lVert}{\rVert}

\usepackage{url}

\def \by{$ \times $}
\def \ninstances{$ R $}
\def \instanceidx{$ r $}
\def \batchsize{$ N $}
\def \patchsize{$ M $}
\def \nclasses{$ C $}
\def \nbands{$ B $}
\def \nkernels{$ K $}
\def \outnkernels{$ K^{\prime} $}
\def \kernelsize{$ G $}
\def \kernelstride{$ S $}
\def \zeropadding{$ Z $}
\def \nrows{$ H $}
\def \ncols{$ W $}
\def \outnrows{$ H^{\prime} $}
\def \outncols{$ W^{\prime} $}
\def \featuremap{$ \mathbf{x} $}
\def \inputpan{$\text{\featuremap}_{PAN}$}
\def \inputms{$\text{\featuremap}_{MS}$}
\def \outfeaturemap{$ \mathbf{x}^{\prime} $}
\def \kernel{$ \mathbf{w} $}
\def \biasvalue{$ \mathrm{b} $}

\def \lossvalue{$ E $}
\def \regularizedloss{$ Q $}
\def \prediction{y}
\def \predictionvector{$ \mathbf{\prediction} $}
\def \predictionvalue{$ \mathrm{\prediction} $}
\def \outscore{$\text{\predictionvector}_{scores}$}
\def \targetvector{$ \mathbf{t} $}

\def \scheduler{$ \eta $}
\def \momentum{$ \alpha $}
\def \weightdecay{$ \lambda $}
\def \nepoch{$ \mathcal{T} $}

\definecolor{myblue}{RGB}{91,155,213}

\begin{document}

\title{Recurrent Multiresolution Convolutional Networks for VHR Image Classification}

\author{John~Ray~Bergado,
        Claudio~Persello,~\IEEEmembership{Senior Member,~IEEE,}
        and~Alfred~Stein%
\thanks{Acknowledgement here.}
\thanks{Manuscript received Month DD, 2017; revised Month DD, 2017.}}

\markboth{IEEE TRANSACTIONS ON GEOSCIENCE AND REMOTE SENSING,~Vol.~X, No.~X, Month~2017}%
{Bergado \MakeLowercase{\textit{et al.}}: Recurrent fully-convolutional networks for VHR image fusion, classification, and map regularization}

\maketitle

\begin{abstract}

Classification of very high resolution (VHR) satellite images has three major challenges: 1) inherent low intra-class and high inter-class spectral similarities, 2) mismatching resolution of available bands, and 3) the need to regularize noisy classification maps. Conventional methods have addressed these challenges by adopting separate stages of image fusion, feature extraction, and post-classification map regularization. These processing stages, however, are not jointly optimizing the classification task at hand. In this study, we propose a single-stage framework embedding the processing stages in a recurrent multiresolution convolutional network trained in an \textit{end-to-end} manner. The feedforward version of the network, called \textit{FuseNet}, aims to match the resolution of the panchromatic and multispectral bands in a VHR image using convolutional layers with corresponding downsampling and upsampling operations. Contextual label information is incorporated into \textit{FuseNet} by means of a recurrent version called \textit{ReuseNet}. We compared \textit{FuseNet} and \textit{ReuseNet} against the use of separate processing steps for both image fusion, e.g. pansharpening and resampling through interpolation, and map regularization such as \textit{conditional random fields}. We carried out our experiments on a land cover classification task using a Worldview-03 image of Quezon City, Philippines and the ISPRS 2D semantic labeling benchmark dataset of Vaihingen, Germany. \textit{FuseNet} and \textit{ReuseNet} surpass the baseline approaches in both quantitative and qualitative results.

\end{abstract}


\IEEEpeerreviewmaketitle

\section{Introduction}

\begin{figure}[tb]
\includegraphics[width=0.5\textwidth]{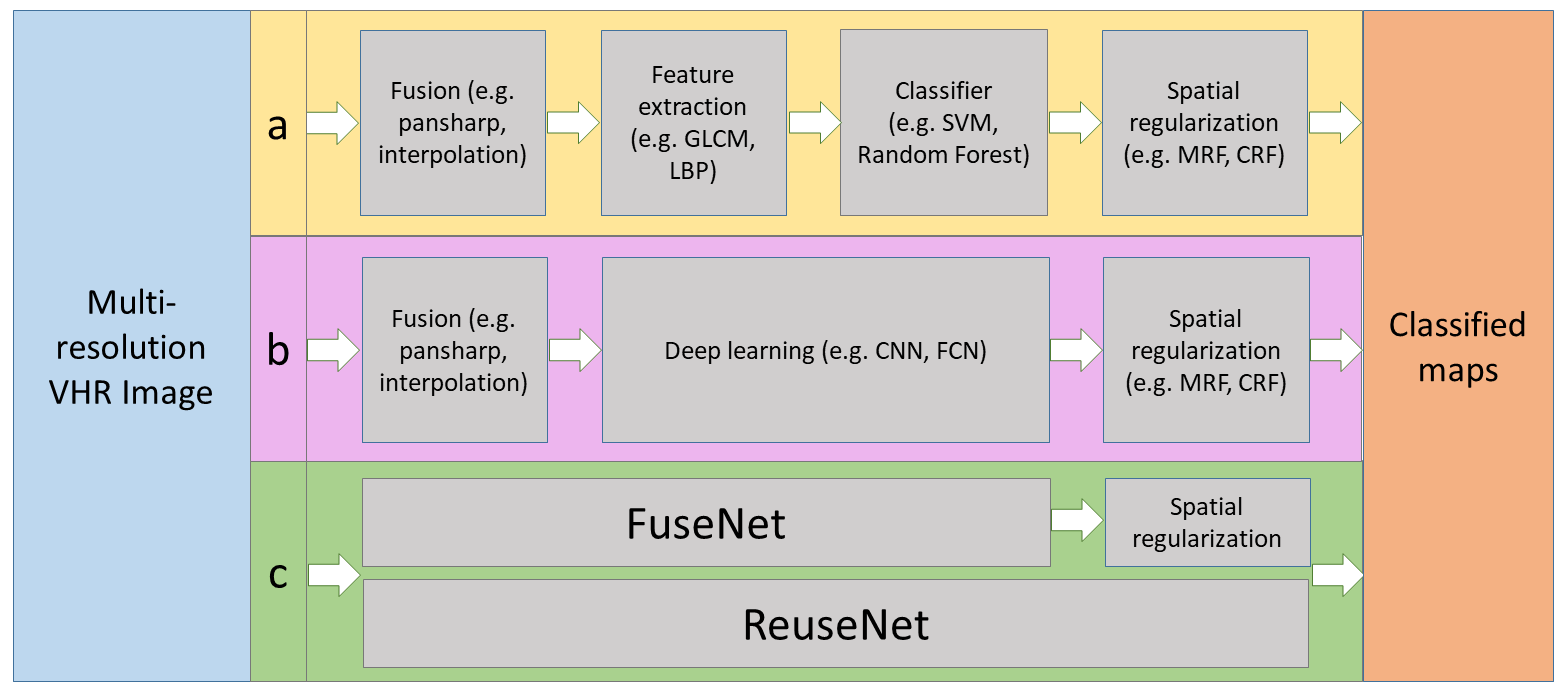}
\caption{Illustration comparing a standard (a), state-of-the-art (b), and proposed (c) piplelines for classifying multiresolution VHR images.}
\label{fig:ReuseNet_pipeline}
\end{figure}

\IEEEPARstart{C}{lassification} of very high resolution (VHR) remotely sensed images allows us to automatically produce maps at a level of detail comparable to conventional in-situ mapping methods. Due to the high spatial resolution of such images, automated classification comes with a set of challenges. One challenge is the inherent low intra-class and high inter-class spectral similarities, inhibiting discrimination of the classes of interest. Conventional methods address this challenge by extracting spatial-contextual features from the image such as texture-describing measures, e.g. gray level co-occurrence matrix (GLCM) \cite{Haralick1973} and local binary patterns (LBP) \cite{Ojala2002}, or products of morphological operators \cite{DallaMura2010,Fauvel2013}. This step is crucial for obtaining discriminative features and accurate classification. However, such feature extraction methods are often disjoint from the supervised classifier, and, hence, not optimized for the task at hand. Deep learning offers a framework to build end-to-end classifiers---directly learning the predictions from the inputs with minimal or no pre-classification and post-classification steps. Features automatically extracted by deep learning based classifiers such as convolutional neural networks (CNN) \cite{LeCun1998} perform better than intermediate handcrafted features \cite{Bergado2016,Mboga2017}. These networks automatically learn spatial-contextual features directly from the input VHR image---effectively integrating the feature extraction step into the training of the classifier as shown in Figure \ref{fig:ReuseNet_pipeline} (b). The design of network architecture, inspired by the model of the visual cortex \cite{Hubel1962}, makes CNN suitable for image analysis and land cover classification.

Another challenge in the classification of VHR images is the multiresolution nature of the images acquired by space-borne sensors. Most VHR satellite images (e.g. Quickbird, Worldview, IKONOS, and Pleiades) capture panchromatic (PAN) images in a spatial resolution four times of the multispectral (MS) bands. Such mismatch in spatial resolution of the images requires an additional step to fuse these images before performing the semantic analysis. Pansharpening and interpolation-based resampling are common techniques for fusing a multiresolution image \cite{Ha2013}. Similar to conventional feature extraction methods, the operations to fuse multiresolution bands of a VHR image add another separate pre-classification step that is disjoint from the training of the supervised classifier, and, hence, may not be optimal for the task at hand. CNN extracts a hierarchy of spatial features at different resolutions. We can exploit the multiresolution nature of the VHR data to design a CNN that performs fusion and feature extraction at the same time.

Literature shows that classification accuracy can be improved by using post-classification spatial regularization \cite{Chen2016,Paisitkriangkrai2016,Wang2017}. Methods employing graphical models, such as conditional random fields (CRF) and Markov random fields (MRF), provide a way to perform this spatial regularization step. Similar to the two pre-classification steps described above, a post-classification map regularization technique adds another step independent of the training of the classifier itself---further including a separate objective function to be optimized. For classifying a multiresolution VHR image, a typical classification pipeline would be composed of three main stages: a pre-classification step performing image fusion and feature extraction, a supervised learning algorithm performing the classification, and a post-classification step regularizing the maps obtained from the supervised classification algorithm. This conventional approach is shown in Figure \ref{fig:ReuseNet_pipeline} (a).

Convolutional networks have been recently applied to classify remotely sensed images with very high resolution \cite{Paisitkriangkrai2016,Sherrah2016,Maggiori2017,Mboga2017,Persello2017,Volpi2017,Zhao2017}. But, aside from \cite{Xu2017} which used a combination of patch-based CNN and stacked autoencoders to fuse PAN and MS images, the majority of the works did not address the problem of multiresolution VHR images. A patch-based CNN \cite{Mboga2017} and a fully convolutional network (FCN) \cite{Persello2017} were used to detect informal settlements from a pansharpened VHR image. Fully convolutional networks were also used to classify urban objects in VHR images both acquired in aerial and space-borne sensors \cite{Paisitkriangkrai2016,Sherrah2016,Maggiori2017,Volpi2017}. Moreover, \cite{Paisitkriangkrai2016,Sherrah2016,Zhao2017} also utilized a separate post-classification step for map regularization. In this paper, we design a novel single-stage network performing image fusion, classification, and map regularization of a multiresolution VHR image in an end-to-end manner.

\subsection{Contributions}

We propose a multiresolution convolutional network, called FuseNet, and its recurrent version, called ReuseNet, to perform image fusion, classification, and map regularization of a multiresolution VHR image in an end-to-end fashion. We summarize the main contributions of this paper in: image fusion, map regularization, and sensitivity analysis of network parameters.

\subsubsection{Image fusion}

We propose a convolutional network learning how to fuse a multiresolution VHR image, extract spatial features, and classify the latter into classes of interest all at the same time. We call this network FuseNet. It uses convolutional layers with corresponding downsampling and upsampling operations to learn to match and fuse the multiresolution channels of a multispectral VHR image.

\subsubsection{Contextual label dependency through network recurrence}

We incorporate recurrence in the FuseNet architecture to model contextual \textit{label-to-label} dependencies and effectively regularize classification maps. We call this improved version ReuseNet. Contextual label dependencies are incorporated in ReuseNet by feeding classification scores of a previous FuseNet instance to a succeeding one. Moreover, we introduce and compare a novel method to initialize the parameters and initial score maps of a ReuseNet.

\subsubsection{Sensitivity analysis}

We analyze the sensitivity of the network to some of its chosen hyperparameters. We investigate the effect of varying a number of hyperparameters of our network to the classification performance. The considered hyperparameters are: the bottleneck feature map dimensions, the number of convolutional layers, the input patch sizes, the upsampling operations, and the number of FuseNet instances within a ReuseNet.

\section{Convolutional Networks}

\subsection{Background}

Convolutional neural networks are a variant of artificial neural networks connected in a sequential feedforward fashion employing convolutional and pooling (aggregational) operations. Convolutional operations greatly reduce the number of learnable parameters and allows the network to use the same filter to detect the same spatial pattern over different parts of an image. Pooling with downsampling enables the network to learn some degree of translational invariance.

Recurrent neural networks are artificial neural networks employing feedback connection, i.e. connections form a directed cycle. For example, the Jordan network \cite{Jordan1997} has connections from the output units back to the hidden units. Two key concepts namely, parameter sharing and graph unfolding \cite[pp. ~369--372]{Goodfellow2016}, allow these networks to accept input sequences of variable lengths while maintaining model complexity---making recurrent networks widely applied to sequential data. However, parameter sharing and graph unfolding can also be used to design networks for different purposes, e.g. application to non-sequential data, while still taking advantage of the benefits, such as model compactness, from the two concepts \cite{Pinheiro2014}.

\subsection{Deep Networks as Data-flow Graphs}

We can generalize any variant of deep networks by seeing them as data-flow graphs---a graph representing how a set of input data are processed along a possibly branching chain of functions, in the end producing a final set of outputs.
Using such a model, we define the networks by three elements: the sets of data they take as an input, the operations they perform in each function block, and the intermediate and final set of outputs they produce.
Aside from these three key elements of data-flow graphs, details of a unique configuration and instance of a convolutional network are defined by its \textit{hyperparameters} and \textit{parameters} respectively. Hyperparameters are associated with the configuration of a network architecture and are set to fixed values before training the network. Parameters are values associated to a specific network instance and are learned during network training.

\subsubsection{Input}

A convolutional network receives as an input either the whole image itself to be classified or a subset of it, called an input patch.
The dimension of this patch is defined by the patch size hyperparameter \patchsize\ and the number of bands \nbands.
A convolutional network accepts an \batchsize \by \nbands \by  \patchsize \by \patchsize\ array of pixel values as an input (in the case of the image patches having equal height and width), \batchsize\ being the number of patches processed by the network in parallel. Aside from the input image patch, the corresponding reference image can also be considered as an input in terms of data-flow graphs since no operation precedes it.

\subsubsection{Operations}\label{subsub:ops}

\textit{Convolutions} are the main operations used by convolutional networks. A convolution applies a linear operation on an input image/feature map using a set of \outnkernels\ learnable kernels.
Applying a kernel \kernel, composed of a \nkernels \by \outnkernels \by \kernelsize \by \kernelsize\ array of learnable parameters, on a \nkernels \by \nrows \by \ncols\ input feature map \featuremap, where \kernelsize\ is the kernel size, \nkernels\ is the number of kernels in each set of kernels, and \nrows\ and \ncols\ are the height and width of the feature map, produces a \outnkernels \by \outnrows \by \outncols\ output feature map \outfeaturemap. The output at the $i^{\prime}$ row and $j^{\prime}$ column of the $k^{\prime}$ feature map is given by:
\begin{linenomath}
\begin{subequations}\label{eq:convolutions}
	\begin{align}
		\text{\outfeaturemap}_{k^{\prime}i^{\prime}j^{\prime}} &= \sum\limits_{k=1}^{\text{\nkernels}}\sum\limits_{p=1}^{\text{\kernelsize}}\sum\limits_{q=1}^{\text{\kernelsize}}\text{\featuremap}_{kij} \cdot \text{\kernel}_{kk^{\prime}pq} + \text{\biasvalue}_{k^{\prime}}\label{eq:convolution}\\
		i &= i^{\prime}+p-\ceil{\frac{\text{\kernelsize}}{2}}\\
		j &= j^{\prime}+q-\ceil{\frac{\text{\kernelsize}}{2}}
	\end{align}
\end{subequations}
\end{linenomath}
where \biasvalue$_{k^{\prime}}$ is the learnable bias parameter associated with the $k^{\prime}$ feature map. The width and height of the output feature map are given by:
\begin{linenomath}
\begin{subequations}\label{eq:outputdim}
	\begin{align}
		\text{\outnrows} &= \floor{\frac{\text{\nrows}-\text{\kernelsize}+2\text{\zeropadding}}{\text{\kernelstride}}+1} \label{eq:outnrows}\\
		\text{\outncols} &= \floor{\frac{\text{\ncols}-\text{\kernelsize}+2\text{\zeropadding}}{\text{\kernelstride}}+1} \label{eq:outncols}
	\end{align}
\end{subequations}
\end{linenomath}
where the \textit{zero-padding} \zeropadding\ is the number of rows and columns of zeros added to the border of the input feature map and the convolutional \textit{stride} \kernelstride\ is the number of units separating contiguous receptive fields of the kernel on the input feature map.

\textit{Nonlinearity} is applied after the linear operation of a convolution. Since applying a series of linear operations can be reduced to a single linear operation, an elementwise nonlinear function applied between each convolution allows the network to learn more complex input to output mapping. A common choice is the rectifier function
\begin{linenomath}
\begin{equation}
		\text{\outfeaturemap}_{i^{\prime}j^{\prime}k^{\prime}} = \max(0, \text{\featuremap}_{ijk})\label{eq:rectifier}
\end{equation}
\end{linenomath}
or a variation of it \cite{Maas2013,He2015b,Clevert2016}.

\textit{Pooling} takes an aggregate of values over local regions of the input. A common choice of a pooling function is the average or maximum function. 
In contrast to convolution, a basic pooling does not have any learnable parameters. Originally, pooling was used to give the network a small degree of translation invariance by summarizing values of the input on non-overlapping windows ($\text{\kernelstride}=\text{\kernelsize}$)---also downsampling the input by a factor of \kernelstride, with proper zero-padding.

\textit{Upsampling} operations are applied to increase the spatial dimensions of input feature maps. Upsampling is important specifically if the network needs to produce output predictions of the same size as the input, i.e. we want to produce a label for each pixel in the \patchsize\by\patchsize\ input patch. One way to upsample is by employing resampling techniques such as nearest neighbor or bilinear interpolation \cite{Chen2016}. The original fully convolutional network (FCN) \cite{Long2015} learns the upsampling operation using \textit{backwards convolution} (or more technically fitting called \textit{transposed convolution}).

\textit{Merging} combines two or more sets of feature maps in a network either by addition or by concatenation. Addition is an elementwise operation performed between feature maps---adding each unit with corresponding indices---hence, all the three dimensions (\nkernels, \nrows, \ncols) must be the same for all inputs \cite{Long2015}. Concatenation stacks the input feature maps depth-wise---hence, only the spatial dimensions (\nrows, \ncols) must be the same.

\subsubsection{Outputs}

In data-flow graph terms, the outputs of a convolutional network consist of all the intermediate feature maps, the final class score maps, and the corresponding loss and accuracy calculated using the class score maps and the reference labels.
Final class score maps correspond to the units in the last layer of a neural network and its dimension depends on how the task is defined. Authors in \cite{Volpi2017} categorize the approaches to this task into three variants: 1) patch classification, 2) subpatch labeling, and 3) full patch labeling. In patch classification, we assign a single label to the patch, i.e. the label corresponds to the class of the central pixel of the patch \cite{Bergado2016,Volpi2017,Mboga2017}. In subpatch labeling, we assign labels on a smaller part of the patch corresponding to the area near the center of the patch \cite{Volpi2017}. Finally, in full patch labeling, we assign labels to all the pixels in the patch \cite{Long2015,Sherrah2016, Badrinarayanan2017,Volpi2017,Persello2017}. The last method, aside from being more efficient, also decouples the limit of the input patch size to the number of downsampling operations in the network.

\subsection{Training Deep Networks}\label{sub:learning}

We train the network by minimizing an objective function in terms of the parameters of the network. For classification involving \nclasses\ classes, a cross-entropy loss function is often used given by:
\begin{linenomath}
\begin{equation}\label{eq:crossentropy}
    \text{\lossvalue}_{\text{\batchsize}}(\text{\kernel}) = -\sum\limits_{n=1}^{\text{\batchsize}} \text{\targetvector}_{n} \cdot \log(\text{\predictionvector}_{n})
\end{equation}
\end{linenomath}
where \lossvalue\ is the loss function value evaluated over \batchsize\ samples, $\text{\targetvector}_{n}$ is a binary vector encoding the the target class labels (with the index corrresponding to a class having a value of 1 and 0 otherwise), $\cdot$ denotes the dot product, and $\text{\predictionvector}_{n}$ is the class score maps of a sample $n$ calculated using a \textit{softmax} activation function:
\begin{linenomath}
\begin{equation}\label{eq:softmax}
    \text{\predictionvector}_{kij} = \frac{\exp(\text{\featuremap}_{kij})}{\sum\limits_{c=1}^{\text{\nclasses}}\exp(\text{\featuremap}_{cij})}.
\end{equation}
\end{linenomath}
In this equation, \predictionvector\ is the softmax score and \featuremap\ is the last set of feature maps containing unnormalized class scores at location $ij$.

A stochastic version of the backpropagation with gradient descent algorithm is often used to minimize the objective function \cite{Bottou2012}. The training is finished after a fixed number of epoch or when a certain convergence criterion is met. We can infer predictions from the final trained network instance by truncating the loss evaluation in the computational graph and taking the index of the maximum class score map value along the class score dimension by
\begin{linenomath}
\begin{equation}\label{eq:inference}
    \text{\predictionvalue}_{ij} = \argmax_{c} \text{\predictionvector}_{cij}
\end{equation}
\end{linenomath}
where $\text{\predictionvector}$ and $\text{\predictionvalue}$ are the class score and prediction for location $ij$ respectively.

\subsection{Regularizing Deep Networks}\label{sub:regularization}

Deep networks are often prone to overfit the training set. Overfitting occurs when a model reports high accuracy during training but performs poorly on unseen test data. Regularization approaches address the overfitting problem using three common methods: \textit{data augmentation}, \textit{weight decay}, and \textit{early-stopping}. Data augmentation technique increases the number of training samples by permuting them with applicable rotational and/or translational transformations. Data augmentation helps the network to learn relevant invariances that may be present in the input. Weight decay modifies the loss function by
\begin{linenomath}
\begin{equation}\label{eq:weightdecay}
    \text{\regularizedloss}(\text{\kernel}) = \text{\lossvalue}(\text{\kernel}) + \text{\weightdecay}\norm{\text{\kernel}}^{2}_{2}
\end{equation}
\end{linenomath}
adding a penalty proportional to the square of the $\mathit{l_2}$-norm of the weight vector \kernel. The weight decay \weightdecay\ hyperparameter controls the contribution of this penalty to the loss function.
Early stopping prematurely stops the training when a criterion measured from a validation set is met.

\section{Proposed Approach}

In this paper, we propose a multiresolution convolutional network, called FuseNet, and its recurrent version, called ReuseNet, to perform an end-to-end fusion, classification, and map regularization of a multi-resolution VHR image. ReuseNet is built on top of a fully convolutional network architecture learning to: 1) fuse PAN and MS bands of a VHR image, 2) perform land cover classification on the fused images, and 3) spatially regularize the resulting classification.

\begin{figure*}[!htb]
\centering
\includegraphics[width=0.9\textwidth]{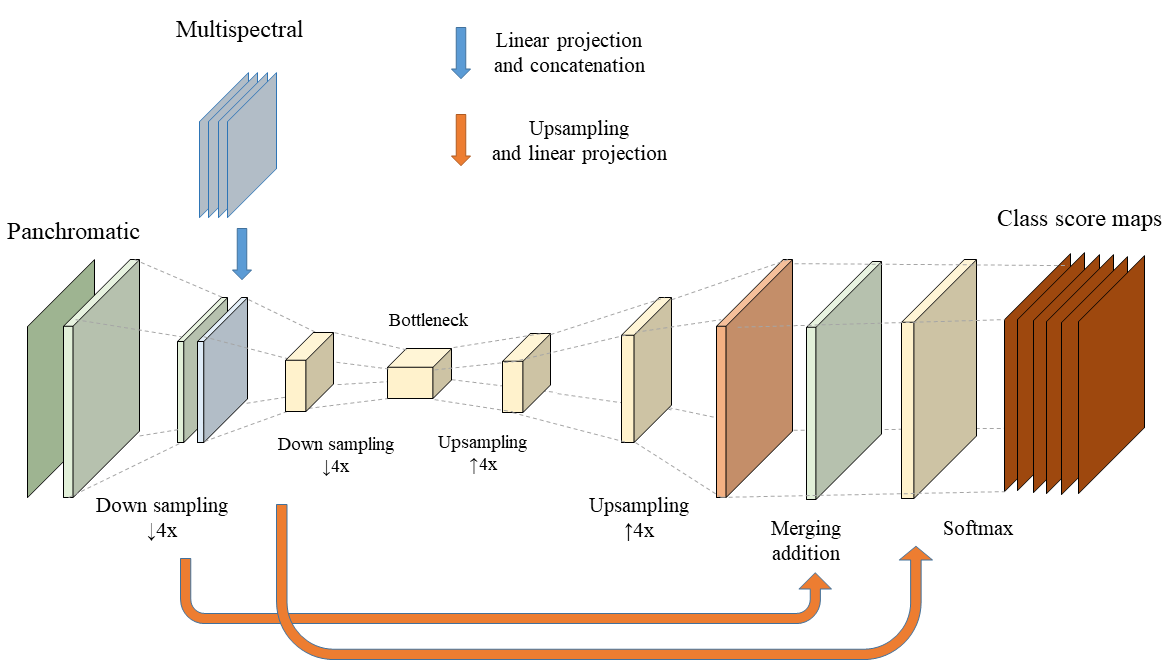}
\caption{The general architecture of FuseNet with skip connections (FuseNet$_{skip}$). FuseNet accepts two streams of input: one from PAN image patches and another from MS image patches. It applies convolutional and pooling layers with downsampling to extract spatial features and at the same time match the resolution of the two streams of input. Similar operations are performed to the output of the merged streams of input arriving at a feature map with smallest spatial dimensions (bottleneck). From there, upsampling operations using transposed convolutions are performed to restore the resolution back to the resolution of the PAN image patches. Skip connections are implemented using appropriate upsampling and linear projections.}
\label{fig:fusenet}
\end{figure*}

\subsection{FuseNet}\label{sub:fusenet}

The architecture of FuseNet is inspired by several encoder-decoder like convolutional network structures \cite{Noh2015,Badrinarayanan2017,Volpi2017} where the first set of layers learn deep features by a series of convolution, nonlinearity, and maximum pooling with downsampling operations, followed by a second set of layers using upsampling and nonlinearity operations to restore the resolution of the original input image. The main difference of FuseNet with these encoder-decoder architectures is the two initial separate streams of the downsampling part of the network that learns how to fuse two images of different resolution. FuseNet is specifically designed for VHR satellite images with PAN band and MS bands of ground sampling distance ratio of four (e.g. Quickbird, Worldview 2/3, Pleiades, Ikonos). But the architecture can be further generalized to fuse any number of images with different spatial resolutions.

FuseNet accepts two sets of input: an image patch \inputpan\ of dimensions \batchsize\by1\by4\patchsize\by4\patchsize\ taken from PAN image and another patch \inputms\ of dimensions \batchsize\by4\by\patchsize\by\patchsize\ taken from the corresponding location in the MS image. It performs two series of convolution, nonlinearity, and maximum pooling with downsampling to \inputpan\ such that the spatial dimensions of the intermediate feature maps match the spatial dimensions of \inputms. The nonlinearity operations use an exponential linear activation function \cite{Clevert2016}. The second input is linearly projected in $k$ dimensions using 1\by1 convolutions such that $k$ matches the number of intermediate feature maps extracted from the first set of input---ensuring a balanced contribution from the two streams of feature. FuseNet then merges the linear projection of \inputms\ with intermediate feature maps extracted from \inputpan\ via a concatenation operation.

Additional series of convolution, nonlinearity, and maximum pooling with downsampling operations are applied to the merged feature maps producing the set of feature maps with smallest spatial dimensions---called \textit{bottleneck} \cite{Volpi2017}. FuseNet then upsamples the bottleneck back to the resolution of \inputpan\ using transposed convolutions. The resulting set of feature maps is linearly projected again using 1\by1 convolutions such that the number of feature maps matches \nclasses. The final class score maps \outscore\ are obtained by applying a softmax activation function. This series of operations can be formulated as a function composition given by:
\begin{linenomath}
\begin{equation}\label{eq:fusenet_ops_long}
    \text{\outscore} = s(l_{1}(u(d_{1}(d_{0}(\text{\inputpan}) \oplus l_{0}(\text{\inputms})))))
\end{equation}
\end{linenomath}
where $d_{i}$ is a series of convolution, nonlinearity, and maximum pooling with downsampling operations, $u$ is the series of upsampling operations via transposed convolution, $l_{i}$ are the linear projections via 1\by1 convolutions, $s$ is the softmax function, and $\oplus$ denotes merging via concatenation. Details of each operation, including the hyperparameter values and dimensions of some chosen intermediate feature maps, are provided in Table \ref{tab:fusenet_ops}. A cross-entropy function following Equation \ref{eq:crossentropy} is used to compute the loss in each iteration. Unlabeled pixels are assigned a loss function value of zero.

We described above the default configuration of FuseNet, called FuseNet$_{low}$, performing the fusion at the lower (MS image) resolution. We also tested a network, called FuseNet$_{skip}$, adding skip connections to some lower-level feature maps of FuseNet$_{low}$ \cite{Long2015}. Figure \ref{fig:fusenet} shows a diagram illustrating the general architecture of FuseNet$_{skip}$. Additionally, we experimented with a FuseNet performing the fusion at the resolution of the PAN image, called FuseNet$_{high}$ which is more similar to pansharpening---upsampling \inputms\ first before fusing them with \inputpan.
Tables \ref{tab:fusenet_ops} and \ref{tab:fusenet_skip_ops} show details of the operations, including dimensions of intermediate output feature maps, used by the FuseNet variants. The format is adapted from \cite{Simonyan2014}. \inputpan\ and \inputms\ denote input patches from the PAN and MS images respectively. IFM and BFM correspond to intermediate and bottleneck feature maps. Convolutions are denoted as ``conv$\langle$kernel size \kernelsize$\rangle$-$\langle$number of kernels \nkernels$\rangle$''. Maximum pooling operations (maxpool) are fixed to have pooling size \kernelsize$_{p}$ and stride \kernelstride$_{p}$ equal to two. Upsampling operations are denoted as ``ups$\langle$number of kernels \nkernels$\rangle$-$\langle$upsampling factor$\rangle$''. Merging operations are denoted as concat for concatenation and add for addition. Fixed upsampling can either be via pansharpening or bilinear interpolation. Consecutive Batch Normalization \cite{Ioffe2015} and exponential linear activation \cite{Clevert2016} operations between convolutions and pooling are omitted for brevity. Finally, operations shared by separate streams of feature align with the columns of these streams.

\begin{table*}[!t]
\centering
\caption{Detailed operations of FuseNet$_{low}$, FuseNet$_{high}$, and FuseNet$_{pansharp/bilinear}$.}
        \begin{threeparttable}
        \begin{tabular}{|c|c|c|c|c|c|}
		\hline
		\multicolumn{2}{|c|}{FuseNet$_{low}$} & \multicolumn{2}{c|}{FuseNet$_{high}$} & \multicolumn{2}{c|}{Net$_{pansharp/bilinear}$} \\
		\hline
		 \inputpan (1\by4\patchsize\by4\patchsize) & \inputms (4\by\patchsize\by\patchsize) & \inputpan (1\by4\patchsize\by4\patchsize) & \inputms (4\by\patchsize\by\patchsize) & \inputpan (1\by4\patchsize\by4\patchsize) & \inputms (4\by\patchsize\by\patchsize) \\
		 \hline
		 conv13-16 & conv1-32 & & ups2-16 & & \\
		 maxpool & & & ups2-8 & & fixed \\
		 conv7-32 & & & conv1-4 & & upsampling \\
		 \cline{4-4}
		 maxpool & & & IFM1 (5\by4\patchsize\by4\patchsize) &  &  \\
		 \cline{3-6}
		  & & \multicolumn{2}{c|}{ concat } & \multicolumn{2}{c|}{ concat } \\
		 \cline{3-6}
		  & & \multicolumn{2}{c|}{ IFM3 (5\by4\patchsize\by4\patchsize) } & \multicolumn{2}{c|}{ IFM3 (4\by4\patchsize\by4\patchsize) } \\
		 \cline{3-6}
		  & & \multicolumn{4}{c|}{ conv13-16 } \\
		 \cline{1-2}
		 IFM1 (32\by\patchsize\by\patchsize) & IFM2 (32\by\patchsize\by\patchsize) & \multicolumn{4}{c|}{maxpool} \\
		 \cline{1-2}
		 \multicolumn{2}{|c|}{concat} & \multicolumn{4}{c|}{conv32-7} \\
		 \cline{1-2}
		 \multicolumn{2}{|c|}{IFM3 (64\by\patchsize\by\patchsize)} & \multicolumn{4}{c|}{maxpool} \\
		 \hline
		 \multicolumn{6}{|c|}{conv3-64} \\
		 \multicolumn{6}{|c|}{maxpool} \\
		 \multicolumn{6}{|c|}{conv3-128} \\
		 \multicolumn{6}{|c|}{maxpool} \\
		 \hline
		 \multicolumn{6}{|c|}{BFM (128\by\patchsize/4\by\patchsize/4)} \\
		 \hline
		 \multicolumn{6}{|c|}{ups2-128} \\
		 \multicolumn{6}{|c|}{ups2-64} \\
		 \multicolumn{6}{|c|}{ups2-32} \\
		 \multicolumn{6}{|c|}{ups2-16} \\
		 \multicolumn{6}{|c|}{conv1-6} \\
		 \hline
		 \multicolumn{6}{|c|}{IFM4 (6\by4\patchsize\by4\patchsize)} \\
		\hline
		\multicolumn{6}{|c|}{softmax} \\
		\hline
        \end{tabular}
        \end{threeparttable}
        \label{tab:fusenet_ops}
\end{table*}

\begin{table*}[!t]
\centering
\caption{Detailed operations of FuseNet$_{skip}$.}
        \begin{threeparttable}
        \begin{tabular}{|>{\centering\arraybackslash}p{4cm}|>{\centering\arraybackslash}p{4cm}|c|c|}
		\hline
		\multicolumn{4}{|c|}{FuseNet$_{skip}$} \\
		\hline
		 \inputpan (1\by4\patchsize\by4\patchsize) & \inputms (4\by\patchsize\by\patchsize) & \multicolumn{2}{c|}{Skip-connected layers} \\
		 \hline
		 conv13-16 & conv1-32 & \multicolumn{2}{c|}{}  \\
		 maxpool &  & \multicolumn{2}{c|}{} \\
		 conv7-32 & & \multicolumn{2}{c|}{} \\
		 maxpool & & \multicolumn{2}{c|}{} \\
		 \cline{1-2}
		 IFM1 (32\by\patchsize\by\patchsize) & IFM2 (32\by\patchsize\by\patchsize) & \multicolumn{2}{c|}{} \\
		 \cline{1-2}
		 \multicolumn{2}{|c|}{concat} & \multicolumn{2}{c|}{} \\
		 \cline{1-2}
		 \multicolumn{2}{|c|}{IFM3 (64\by\patchsize\by\patchsize)} & \multicolumn{2}{c|}{} \\
		 \cline{1-2}
		 \multicolumn{2}{|c|}{conv3-64} & \multicolumn{2}{c|}{} \\
		 \multicolumn{2}{|c|}{maxpool} & \multicolumn{2}{c|}{} \\
		 \multicolumn{2}{|c|}{conv3-128} & \multicolumn{2}{c|}{} \\
		 \multicolumn{2}{|c|}{maxpool} & \multicolumn{2}{c|}{} \\
		 \cline{1-2}
		 \multicolumn{2}{|c|}{BFM (128\by\patchsize/4\by\patchsize/4)} & \multicolumn{2}{c|}{} \\
		 \cline{1-2}
		 \multicolumn{2}{|c|}{ups2-128} & \multicolumn{2}{c|}{} \\
		 \multicolumn{2}{|c|}{ups2-64} & \multicolumn{2}{c|}{} \\
		 \multicolumn{2}{|c|}{ups2-32} & \multicolumn{2}{c|}{} \\
		 \cline{3-4}
		 \multicolumn{2}{|c|}{ups2-16} & IFM1 & IFM5 \\
		 \cline{3-4}
		 \multicolumn{2}{|c|}{conv1-6} & ups6-4 & ups6-8 \\
		 \hline
		 \multicolumn{2}{|c|}{IFM6 (6\by4\patchsize\by4\patchsize)} & IFM7 (6\by4\patchsize\by4\patchsize) & IFM8 (6\by4\patchsize\by4\patchsize) \\
		 \hline
		 \multicolumn{4}{|c|}{add} \\
		 \hline
		 \multicolumn{4}{|c|}{IFM4 (6\by4\patchsize\by4\patchsize)} \\
		\hline
		\multicolumn{4}{|c|}{softmax} \\
		\hline
        \end{tabular}
        \end{threeparttable}
        \label{tab:fusenet_skip_ops}
\end{table*}

FuseNet implements a full patch labeling approach since it produces labeled image patches of the same dimensions as the input PAN image patch. Inference of final classification map is given by Equation \ref{eq:inference} and can be applied to an input image of variable spatial dimension. Application to input of variable size is made possible by the fully-convolutional nature of the network \cite{Long2015}---allowing it to be applied as an image filter \cite{Sherrah2016} to any input with spatial dimensions of at least equal to the FCN's effective receptive field.

\subsection{ReuseNet}\label{sub:reusenet}

\begin{figure*}[tb]
\centering
\includegraphics[width=0.9\textwidth]{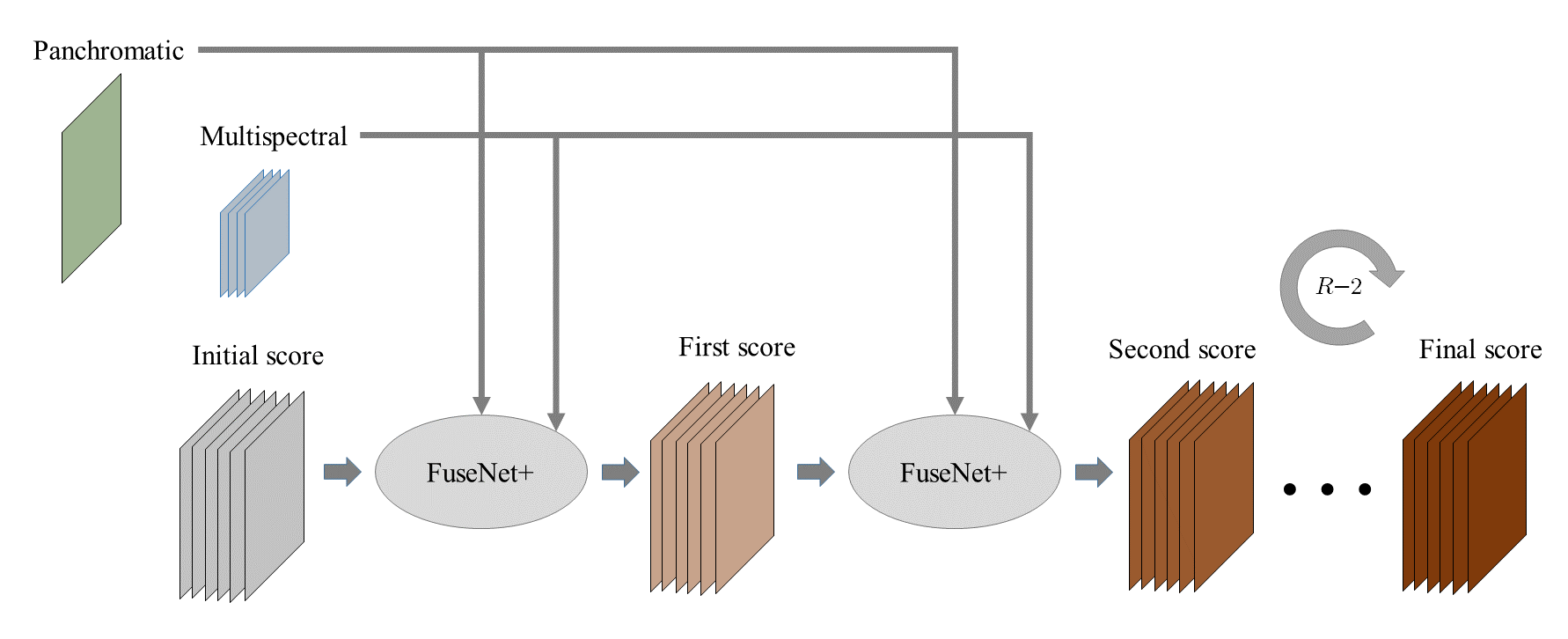}
\caption{The general architecture of ReuseNet with \ninstances\ FuseNet+ instances. FuseNet+ employs exactly the same operations as FuseNet except for the first layer applying additional sets of convolutional filters on the input score maps. ReuseNet accepts three streams of input: 1) \inputpan, 2) \inputms, and 3) score maps of the same resolution as \inputpan. It applies the same operations employed by a FuseNet in \ninstances\ cycles, taking the output score map of the previous cycle as an input.}
\label{fig:reusenet}
\end{figure*}

ReuseNet builds on top of the architecture of FuseNet by incorporating recurrent connections. Incorporation of this recurrent architecture in a full patch labeling approach enables the network to learn contextual label-to-label dependencies by feeding output score maps of a FuseNet instance to another instance of itself as an input. Such dependencies are similar to what graphical model (e.g. CRF/MRF) based methods learn in a post-classification regularization inference. For instance, a fully-connected CRF \cite{Krahenbuhl2011} solves an energy function that penalizes label configurations based on a unary term, often taken as the negative logarithm of the class scores \cite{Chen2016}, and a pairwise term, adding a penalization for pixels with different labels based on image-space and feature-space distances.
For ease of notation, let the series of operations performed by FuseNet (Equation \ref{eq:fusenet_ops_long}) be given by the function $f$:
\begin{equation}\label{eq:fusenet_ops}
    \text{\predictionvector} = f(\text{\inputpan}, \text{\inputms})
\end{equation}
where the \featuremap's are the input of FuseNet, and \predictionvector\ is the class score map resulting from this input. The operations performed by ReuseNet are given by a recurrent variant $g$:
\begin{subequations}\label{eq:reusenet_ops}
	\begin{align}
		\text{\predictionvector}_{1} &= g(\text{\inputpan}\oplus \text{\predictionvector}_{0},\text{\inputms})\\
    \text{\predictionvector}_{\text{\instanceidx}} &= g(\text{\inputpan}\oplus \text{\predictionvector}_{\text{\instanceidx}-1},\text{\inputms})
	\end{align}
\end{subequations}
where the \instanceidx\ score map is obtained by applying the same function to a combination of the previous $\text{\instanceidx}-1$ score map and the original FuseNet input as a new input. The recurrent variant $g$ (denoted as FuseNet+ in Figure \ref{fig:reusenet}) applies exactly the same operations as $f$ except for the first operation that instead of only taking \inputpan\ as an input, this operation takes the concatenation of \inputpan\ and a class score map $\text{\predictionvector}_{r}$ associated to the network instance $r$. Figure \ref{fig:reusenet} shows a diagram illustrating the general architecture of ReuseNet.

We tested ReuseNet with several number of FuseNet instances (2, 3, and 4), calling each ReuseNet-\ninstances\ where \ninstances\ is the number of FuseNet instances within the ReuseNet. We also investigated various methods for initializing weights and initial score maps $\text{\predictionvector}_{0}$ of ReuseNet. Plain ReuseNet initializes the score maps with zeros, while ReuseNet$_{map-init}$ initializes the score maps using scores from a pre-trained FuseNet showing the best results in the fusion comparison experiments. We further extend ReuseNet$_{map-init}$ by initializing the weights of the FuseNet instance in the ReuseNet with the same FuseNet that provides the initial score maps. We call this extension ReuseNet$_{map-weights-init}$.

\subsection{Perspective on Learning the Fusion Approach and Incorporating Recurrence}\label{sub:reusenet_perpsective}

Conventional approaches to classify multiresolution images require a separate step to match the resolution of the images. One way is to spatially sharpen the MS images using the PAN image (also called pansharpening) \cite{Hester2008}. Another possible way is to resample images to match a specific resolution using nearest neighbor or bilinear interpolation. However, these standard fusion techniques are performed independently from the classification problem and are suboptimal.
FuseNet provides a streamlined approach including the fusion of the multiresolution images within the learning of the classifier.
We expect that coupling and learning the fusion method within a supervised classifier will outperform an approach based on a separate fusion method.

The parameter sharing across FuseNet instances in a ReuseNet is consistent with the definition of a recurrent network, i.e. a recurrent network is a feedforward network that keeps on reusing the same set of weights to cycle through a sequence. The authors in \cite{Pinheiro2014} view such incorporation of recurrence as a way to increase the contextual window size, equivalent to the patch size \patchsize\ in a patch classification approach, of their patch classification based approach while controlling the capacity of the network via inter-instance weight sharing. While both increase in contextual window size and capacity control of a CNN-based image patch classifier helps to improve the latter's performance, the first benefit is lost in a full patch labeling approach. In a fully convolutional network implementing full patch labeling, the contextual window size does not change as recurrent operations are added to the network since the contextual window size is equivalent to the effective size of the receptive field of the network. The effective size of the receptive field of the network depends on kernel sizes and strides of the network's convolutional and pooling operations, which are fixed and the same across instances.

In the proposed ReuseNet, recurrence integrates \textit{contextual label information} to our model by considering class score maps as inputs to each FuseNet instance. This allows the model to learn label-to-label dependencies in addition to the spatial contextual information learned from the pixel values, \textit{pixel-to-label} dependencies. This is a form of structured output prediction \cite{Bakir2007} where interdependencies between outputs may be expressed in terms of constraints restricting permissible output combinations or a more flexible form such as spatial dependencies across different output variables. Graphical models such as conditional random fields \cite{Lafferty2001} are commonly used for such structured prediction tasks.
ReuseNet uses operations in a deep convolutional network to learn features from both the input image and class scores---integrating the learning of label-to-label dependencies from the data instead of explicit image-space and feature-space distances as represented in a pairwise potential of CRF. This allows ReuseNet to be trained end-to-end as opposed to a two-stage approach applying a post-classification MRF/CRF as done in \cite{Giorgi2014} and \cite{Sherrah2016}.

\section{Data and Experimental Setup}

\begin{figure}[tb]
\includegraphics[width=0.5\textwidth]{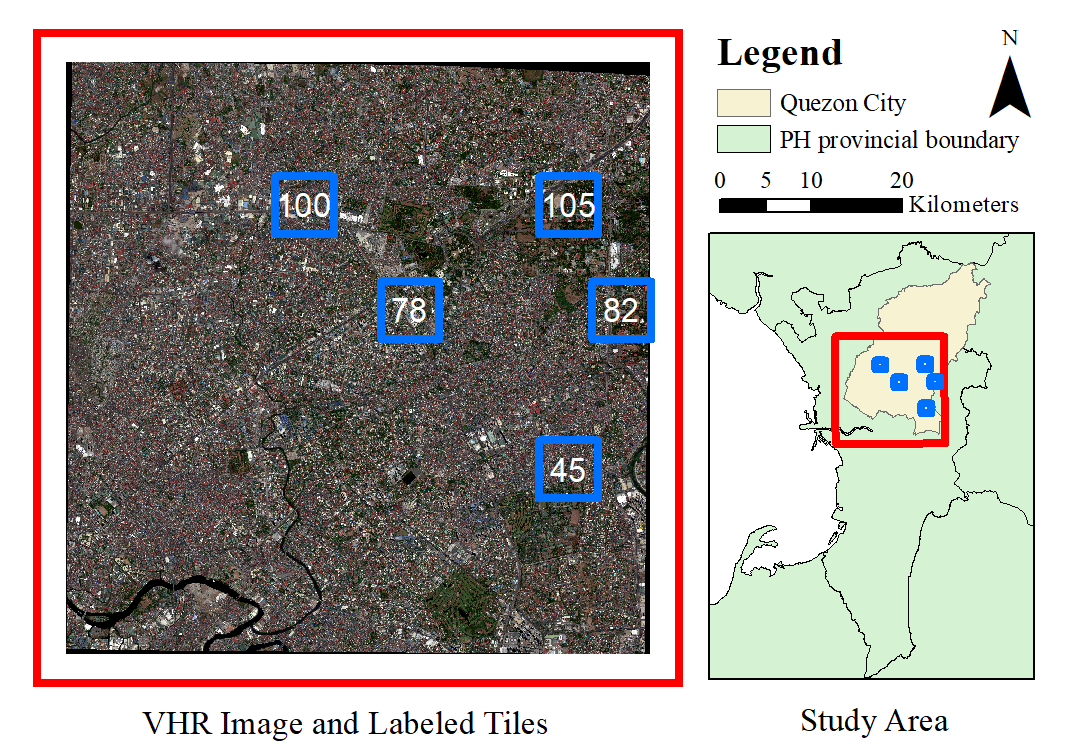}
\caption{Figure showing the true color VHR image together with the locations of the labeled tiles (in blue squares) and the study area: Quezon City, Philippines.}
\label{fig:study_area}
\end{figure}

\subsection{Dataset Description}

\begin{table}[!t]
\centering
\caption{Number of labeled pixels in each tile}
        \begin{threeparttable}
        \begin{tabular}{lcl}
		\hline
		\hline
		Tile		 		& Number of labeled pixels & Set \\
		\hline
		 100 & 2178768 & Training \\
		 105 & 2173602 & Training \\
		 45 & 2063971 & Validation \\
		 78 & 1977336 & Test \\
		 82 & 1961955 & Test \\
		\hline
		\hline
        \end{tabular}
        \end{threeparttable}
        \label{tab:tile_samples}
\end{table}

\subsubsection{Worldview-03 Quezon City dataset} we evaluated the proposed networks in the land cover classification of a dataset covering Quezon City, Philippines. The dataset is composed of a Worldview-03 satellite image of the city acquired on 17$^{th}$ April 2016 and corresponding manually prepared reference images for five chosen tiles (subsets) of the satellite image. The satellite image has a PAN band of 0.3 m resolution and four MS bands (near-infrared, red, green, and blue) of 1.2 m resolution. Reference images were prepared via photointerpretation and set to have the same spatial resolution as the PAN image. The whole satellite image was first divided into regularly-sized image tiles. PAN image tiles have a dimension of 3200 pixels $\times$ 3200 pixels, while MS image tiles have a dimension of 800 pixels $\times$ 800 pixels. Five non-adjacent tiles were sparsely labeled---annotating a pixel with a label belonging to one of the following six classes: impervious surface, building, low vegetation, tree, car, and clutter. Two of the five labeled tiles were used for training (100 and 105), one for validation (45), and the remaining two for testing (78 and 82). Training samples are composed of pairs of image patches with dimensions \patchsize\by\patchsize\ (taken from the MS image) and 4\patchsize\by4\patchsize\ (taken from the PAN image tile).
Figure \ref{fig:study_area} shows the VHR image and the corresponding locations of labeled tiles in the study area while Table \ref{tab:tile_samples} shows the number of labeled pixels in each image tile. Training samples were normalized to have a value between zero and one. The reference image patches have been converted into a ``one-hot'' encoding---a vector having zero values except for the index corresponding to the code of the class.

\subsubsection{ISPRS Vaihingen dataset} for the ReuseNet experiments, we utilized the ISPRS 2D semantic labeling benchmark dataset of Vaihingen as an additional dataset \cite{Cramer2010}. We adopted the experimental setup used in \cite{Sherrah2016,Volpi2017}, employing the same training and validation tiles, to provide comparable results. We followed the sampling done in \cite{Sherrah2016}, except that data augmentation was not applied---resulting in less training samples. The method discussed in \cite{Mousa2017} was employed to extract the normalized DSM.

\subsection{Comparison of methods}\label{sub:methods_comparison}

For the image fusion part, we compared FuseNet against two other baseline approaches: one using pansherpening and another using bilinear interpolation to match the resolution of \inputms\ to the resolution of \inputpan. We call these two baseline approaches Net$_{pansharp}$ and Net$_{bilinear}$. Net$_{pansharp}$ applies Gram-Schmidt pansharpening technique \cite{Laben2000}. Only the pansharpened image is fed as an input to Net$_{pansharp}$. Net$_{bilinear}$ upsamples the resolution of the MS image to match the resolution of the PAN image using bilinear interpolation. The upsampled MS images are then merged to the PAN image using concatenation. The architecture of the network after the fusion is kept the same to have a fair comparison among the different approaches (see details of the FuseNet variants in Table \ref{tab:fusenet_ops}). Additionally, we compared a SegNet \cite{Badrinarayanan2017} trained on the first three principal components of the pansharpened image, since SegNet only accepts three inputs. We found that discarding one band (NIR) considerably degrades the results.

We compared ReuseNet against FuseNet using fully-connected CRF \cite{Krahenbuhl2011} (FuseNet+CRF) to assess the capability of our classifiers to spatially regularize the classification results. The FuseNet+CRF baseline is similar to the approach adopted in \cite{Chen2016,Sherrah2016} but applied to PAN and MS images with different spatial resolutions. Spatial and feature space distances in the pairwise potentials of the fully-connected CRF are computed from the PAN image. We performed a grid-search of the CRF parameters, i.e. the weights and standard deviations of the appearance and smoothness kernels, and used the set of the parameters with the highest accuracy on the validation tile. We fixed the number of iterations to 10 for the mean field approximation algorithm used to perform inference in a fully-connected CRF.

We also performed a sensitivity analysis of a few chosen hyperparameters of FuseNet$_{low}$. We varied the bottleneck feature map dimensions, number of convolutional layers (in the downsampling part of the network), input patch sizes, and upsampling methods---performing the experiments in this order. We took the hyperparameter value that maximizes the overall accuracy on the validation tile and fix it for the succeeding sets of experiments. We experimented using bottleneck feature map dimensions: $16\times16$, $8\times8$, $4\times4$, $2\times2$, and $1\times1$. After fixing the bottleneck feature map dimension, we increased the number of convolutional layers preceding the last downsampling operation---effectively increasing the number of convolutional layers from 8 to 14 in steps of two. We investigated varying patch sizes of $(4\text{\patchsize}, \text{\patchsize})$: $(32, 8)$, $(64, 16)$, $(96, 24)$, $(128, 32)$. For the upsampling operations, we explored two additional methods using nearest neighbor and bilinear interpolation to upsample the feature map and then performing $3\times3$ convolutions after each upsampling operation.

We trained all the networks using a set of 17409 image patches taken from the training tiles and used 8255 image patches taken from the validation tile for early-stopping. We performed a random sampling with the constraint that the pixel near the center of the image patch is labeled. This may produce overlapping patches unlike the systematic gridwise sampling approach used in \cite{Sherrah2016}. Gridwise sampling reduces the number of training patches since the reference images is sparsely labeled, only around five percent of the pixels are labeled.
The total loss value computed over a mini-batch is the total loss of all pixels divided by the number of labeled pixels within the mini-batch.

The FuseNets are trained using backpropagation with stochastic gradient descent setting the initial learning rate $\text{\scheduler}=0.01$, momentum $\text{\momentum}=0.9$, mini-batch size $\text{\batchsize}=32$, and maximum number of epochs $\text{\nepoch}=240$. We decrease the learning rate in a stepwise manner as done in \cite{He2016}---multiplying it by a factor of 0.1 after 60 and 180 epochs. The weights were initialized as in \cite{Glorot2010}. We did not find dropout to be helpful; hence, we only used an $\mathit{l_2}$-weight decay penalty---setting $\text{\weightdecay}=0.001$---and a variant of early-stopping to regularize FuseNet. For early stopping, the classification accuracy on the validation set is calculated every epoch and the last model with the best validation accuracy is fixed to be the final instance of the model.

The FuseNet instances within a ReuseNet are identical, sharing the same network configuration and parameters. Each instance also couples a cross-entropy loss function with each of their score map. The total objective loss of a ReuseNet is the average of the cross-entropy loss values from all the FuseNet instances. We also used the same backpropagation with stochastic gradient descent setting as training a FuseNet with the initial learning rate $\text{\scheduler}=0.01$, momentum $\text{\momentum}=0.9$, mini-batch size $\text{\batchsize}=32$, and maximum number of epochs $\text{\nepoch}=240$. Likewise, we decreased the learning rate in a stepwise manner---multiplying it by a factor of 0.1 after 60 and 180 epochs. For regularization, we only used an $\mathit{l_2}$-weight decay penalty---setting $\text{\weightdecay}=0.001$. We can infer classification map from a ReuseNet in the same manner of inference as a FuseNet, with one additional option: to extract different predictions from each FuseNet instance.

For applying ReuseNet on the ISPRS Vaihingen dataset, we employed a feedforward network similar to the No-downsampling FCN proposed by \cite{Sherrah2016} truncating the last two layers (fc5 and fc6) before softmax activation and entirely removing all maximum pooling without downsampling operations. With only convolutional layers (without pooling), we call this network AllConvNet. The network was trained on 12717 training patches as opposed to the 123330 training patches in \cite{Sherrah2016}. Although having less parameters and having trained with a smaller number of training samples, AllConvNet provided comparable results with the original No-downsampling FCN while requiring less operations. We trained AllConvNet for 150000 iterations as reported in \cite{Sherrah2016}. ReuseNet versions of AllConvNet were applied to the ISPRS Vaihingen dataset and were compared to the best results of both \cite{Sherrah2016} and \cite{Volpi2017}. All the networks in this additional set of experiments were trained using a variant of SGD proposed in \cite{Zeiler2012}.

\subsection{Accuracy Assessment}

We compared the results of the different approaches using global measures: 1) overall classification accuracy (OA), 2) the Kappa coefficient ($\kappa$), 3) average class accuracy (AA), 4) and average class-F1 scores (F1). OA is given by:
\begin{linenomath}
\begin{equation}\label{eq:OA}
    OA = \frac{\sum\limits_{i=1}^{C}n_{ii}}{n}
\end{equation}
\end{linenomath}
where $n_{ii}$ is the number of samples classified as class $i$ in both the the predictions and reference images, $n$ is the total number of labeled samples in the reference images, and $C$ is the number of classes, whereas $\kappa$ is given by:
\begin{linenomath}
\begin{equation}\label{eq:Kappa}
    \kappa = \frac{
                   n\sum\limits_{i=1}^{C}n_{ii} - \sum\limits_{i=1}^{C}n_{i+}n_{+i}
                  }
                  {
                   n^{2} - \sum\limits_{i=1}^{C}n_{i+}n_{+i}
                  }
\end{equation}
\end{linenomath}
where $n_{i+}$ and $n_{+i}$ are the number of samples classified as class $i$ in the predictions and reference images respectively. Both OA and $\kappa$ provides the rate of correctly classified pixels with the latter compensating for random agreement in classification. These global measures, however, are biased toward frequently occurring classes---meaning, classes with less frequencies have relatively little impact to the two measures. Unlike OA and $\kappa$, AA and F1 provides average of measures independent of class distribution. AA is given by:
\begin{linenomath}
\begin{equation}\label{eq:AA}
    AA = \frac{1}{C}\sum\limits_{i=1}^{C}
                    \frac{n_{ii}}
                         {n_{i+}}
\end{equation}
\end{linenomath}
while F1 is given by:
\begin{linenomath}
\begin{equation}\label{eq:F1}
    F1 = \frac{1}{C}
            \sum\limits_{i=1}^{C}
              \frac{
                    2\frac{n_{ii}}
                          {n_{i+}}
                     \frac{n_{ii}}
                          {n_{+i}}
              }{\frac{n_{ii}}
                     {n_{i+}}
                +
                \frac{n_{ii}}
                     {n_{+i}}}
\end{equation}
\end{linenomath}
AA computes the average within-class rate of correctly classified pixels, while F1 calculates the harmonic mean of the precision (user's accuracy) and recall (producer's accuracy). We also observe and comment on the quality of the resulting classified maps.

\section{Results and Discussion}

\subsection{FuseNet}

\begin{table}[!t]
\centering
\caption{Comparison of fusion approaches}
        \begin{threeparttable}
        \begin{tabular}{lcccc}
		\hline
		\hline
		Network		 		& OA (\%) & $\kappa$ (\%) & AA (\%) & F1 (\%) \\
		\hline
		 Net$_{bilinear}$ & 84.76 & 78.70 & 81.99 & 77.48 \\
		 Net$_{pansharp}$ & 86.87 & 81.53 & 82.76 & 77.86 \\
		 SegNet \cite{Badrinarayanan2017} & 88.11 & 83.17 & 83.96 & 77.01 \\
		 FuseNet$_{high}$ & 88.03 & 83.18 & 89.79 & 79.06 \\
		 FuseNet$_{low}$ & 91.63 & 88.03 & 92.91 & \textbf{82.90} \\
		 FuseNet$_{skip}$ & \textbf{91.90} & \textbf{88.43} & \textbf{93.46} & 81.74 \\
		\hline
		\hline
        \end{tabular}
        \end{threeparttable}
        \label{tab:fusion_comparison}
\end{table}

\begin{figure}[tb]
\includegraphics[width=0.5\textwidth]{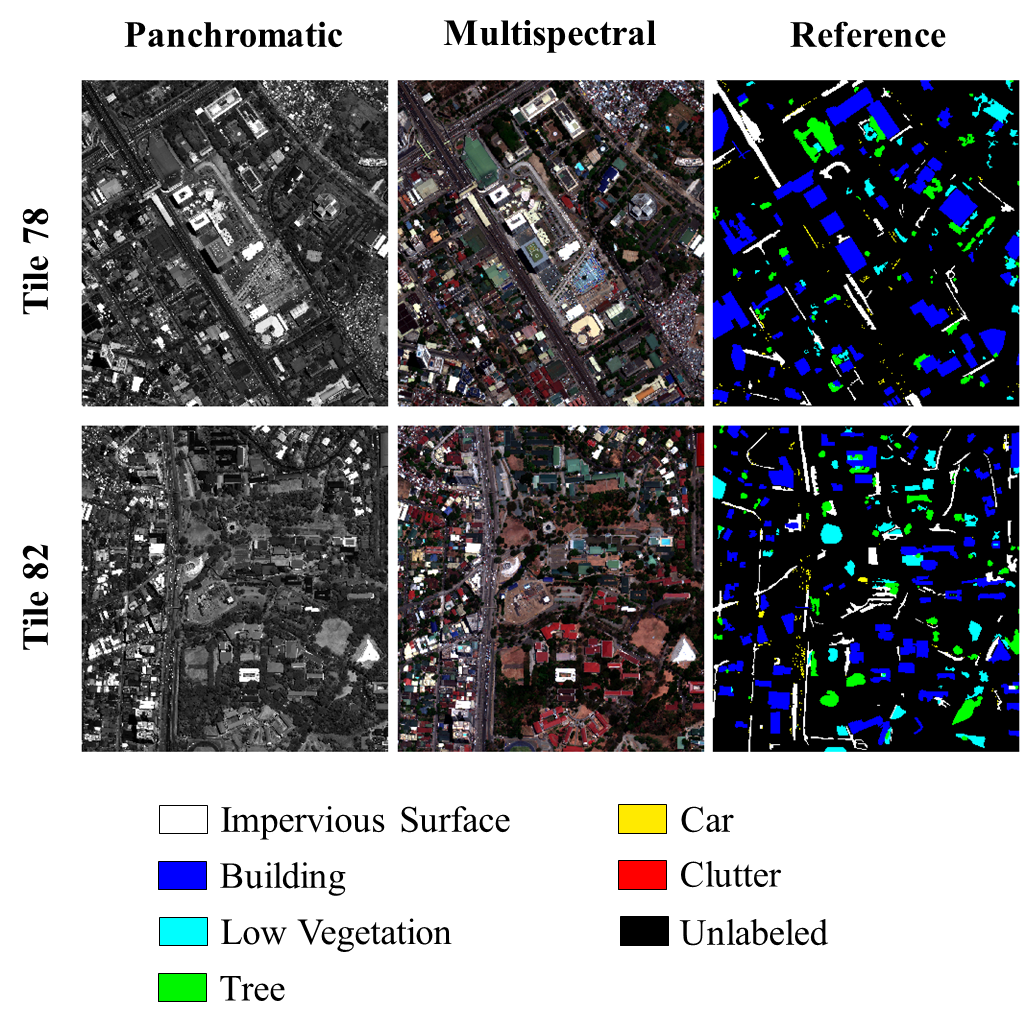}
\caption{PAN, MS, and reference images in the tiles used for testing. Corresponding legend is shown.}
\label{fig:test_tiles}
\end{figure}

\begin{figure*}[tb]
\centering
\includegraphics[width=.8\textwidth]{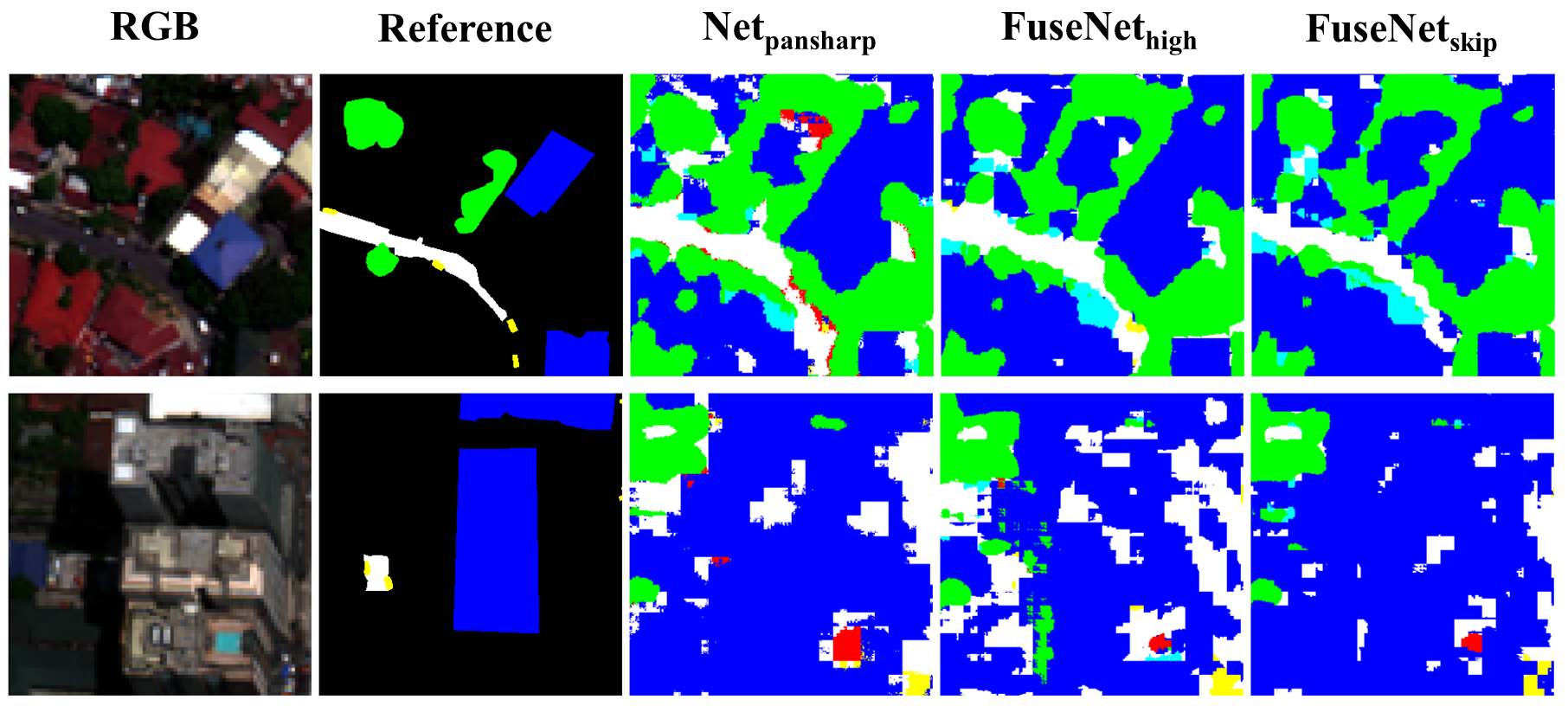}
\caption{Two subsets from the test tiles showing, from right to left, the satellite image (natural color), reference image, and classification maps from selected two FuseNet variants (FuseNet$_{high}$ and FuseNet$_{skip}$) and one baseline method (Net$_{pansharp}$).}
\label{fig:fusenet_plots}
\end{figure*}

Table \ref{tab:fusion_comparison} shows the results of accuracies comparing different fusion approaches. The numerical results are evaluated using all the labeled pixels in the two test tiles (see Table \ref{tab:tile_samples} for the total number test samples). FuseNet$_{skip}$ scores the highest in all the four numerical metrics, except for F1 where FuseNet$_{low}$ scores the highest. FuseNet$_{low}$ outperforms both the variants using fixed upsampling (Net$_{pansharp}$, SegNet, and Net$_{bilinear}$) and the variants learning the upsampling but fusing at the scale of the image with higher resolution (FuseNet$_{high}$). Observing each metric: FuseNet$_{low}$ gains about 3--6\% in OA, 4--9\% in $\kappa$, 3--10\% in AA, and 1--5\% in F1 against the other baselines (with the exemption of FuseNet$_{skip}$). FuseNet$_{skip}$ further increases the numerical results of FuseNet$_{low}$ in the first three metrics by about 0.2--0.5\% but degrades the F1 by about 1.2\%.

We have two relevant observations: 1) learning the fusion can improve the classification of PAN and MS VHR images with different resolutions; 2) fusing at the scale of the image with lower resolution results in better classification than performing the fusion at the scale of the image with higher resolution. The first point demonstrates our expected effectiveness of coupling and learning the fusion operation within a supervised classifier.
One explanation for the second point could be the placement of upsampling layers. Introducing upsampling layers early in the network---as done in FuseNet$_{high}$---may produce artifacts that can degrade its performance.

Figure \ref{fig:test_tiles} shows the PAN, MS, and reference images of the tiles used for testing. Figure \ref{fig:fusenet_plots} shows the classification results of two FuseNet variants (FuseNet$_{high}$ and FuseNet$_{skip}$) and one baseline method (Net$_{pansharp}$) on two selected areas of the test tiles. The most noticeable misclassifications are found in large and high-rise buildings, in both test tiles, and an overpassing road in tile 78. The facades and rooftops of the buildings are often mistaken to be impervious surfaces by the classifiers; while the overpassing road is mistaken to be a building. These regions can appear to have similar spectral characteristics and can only be distinguished by presence of other cues such as appearing to be elevated. However, with the absence of elevation information, such cues are not directly incorporated in the input data. Manually distinguishing arguably vaguely-defined classes such as low-vegetation and impervious surface can also be problematic, especially in the PAN image, with the lack of ancillary information such as elevation. Adding a digital elevation model or a digital surface model can help address the misclassification of these regions. The cars are also generally misclassified by all the classifiers which is, aside from being underrepresented in terms of the number of labeled pixels, due to the lack of spatial resolution of the MS bands and the cars' spectral similarity with other classes (such as impervious surface and buildings) in the PAN band. Overall, FuseNet$_{skip}$ generally has less errors in the facade of large buildings, lessen the artifacts noticeably present in the other techniques, and has better delineation of classes with irregular boundaries such as trees and low-vegetation---providing the best classification results among all the FuseNet variants. We, therefore, apply recurrence to FuseNet$_{skip}$ architecure to build the ReuseNet instances.

\subsection{ReuseNet}\label{sub:reusenet_results}

\subsubsection{Worldview-03 Quezon City dataset}

\begin{table}[!t]
\centering
\caption{Comparison of map regularization approaches on Worldview-03 Quezon City dataset}
        \begin{threeparttable}
        \begin{tabular}{lcccc}
		\hline
		\hline
		Network			 		& OA (\%) & $\kappa$ (\%) & AA (\%) & F1 (\%) \\
		\hline
		 FuseNet & 91.90 & 88.43 & 93.46 & 81.74 \\
		 FuseNet+CRF & 93.07 & 90.08 & \textbf{94.71} & 81.72 \\
		 ReuseNet-2 & 92.82 & 89.69 & 94.09 & 82.64 \\
		 ReuseNet-3 & 92.98 & 89.88 & 94.54 & 85.42 \\
		 ReuseNet-4 & \textbf{93.49} & \textbf{90.58} & 94.53 & 86.67 \\
		 ReuseNet-5 & 92.74 & 89.53 & 92.78 & \textbf{87.29} \\
		\hline
		\hline
        \end{tabular}
        \end{threeparttable}
        \label{tab:regularization_comparison}
\end{table}

\begin{figure*}[tb]
\centering
\includegraphics[width=0.8\textwidth]{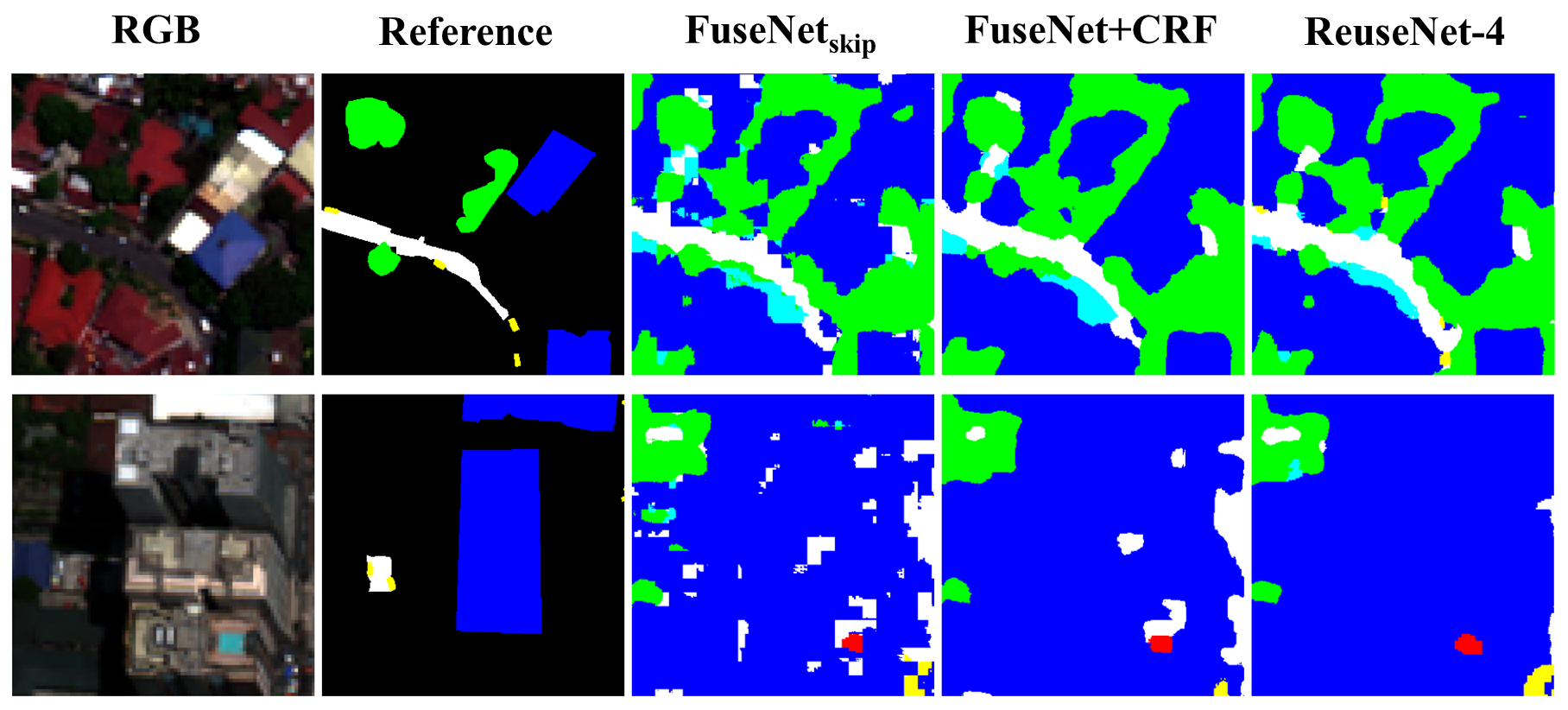}
\caption{Two subsets from the test tiles showing, from right to left, the satellite image (natural color), reference image, and classification maps from FuseNet$_{skip}$, FuseNet$_{skip}$+CRF, and ReuseNet-4. All ReuseNets reported are ``plain'' meaning initial score maps are filled with zeros. Reusenet-\ninstances\ denotes a ReuseNet composed of \ninstances\ number of FuseNet instances.}
\label{fig:reusenet_plots}
\end{figure*}

Table \ref{tab:regularization_comparison} shows the accuracies obtained by comparing different classification techniques on the Worldview-03 Quezon City dataset. We found that both the ReuseNet instances and the baseline method FuseNet+CRF improves the numerical results of the plain FuseNet$_{skip}$ gaining around: 0.9--1.5\% in OA, 1.2--2.1\% in $\kappa$, and 0.6--1.2\% in AA. For the F1, however, FuseNet+CRF method performs worse than the plain FuseNet losing 0.02\%; while all the other ReuseNet instances improves the F1 by around 0.9--5.5\%. ReuseNet-4 outperforms all the other classifiers in all the metrics except for AA and F1---where both ReuseNet-3 and FuseNet+CRF outperform it by some margin in AA (0.01\% and 0.18\% respectively) and ReuseNet-5 considerably outperforms it in F1 by 0.62\%. In particular, all the ReuseNets consistently show better F1 compared to both FuseNet and FuseNet+CRF---gaining almost 6\%. These expected relatively smaller gains in numerical accuracy is consistent with what the author in \cite{Sherrah2016} found---applying a post-classification CRF to an FCN to classify extremely high resolution aerial imagery increases the overall classification accuracy by around 0.1--1.0\%. More noticeable changes are expected in the resulting improved regularity of the classified maps.

The numerical results above supports our assertion that introducing contextual label information through recurrence in an FCN applying a full-patch labeling approach can improve the classification of a VHR image. Such incorporation of label information allows our classifier to learn both pixel-to-label and label-to-label contextual dependencies. We can develop an intuition of these two dependencies by using an analogy to photointerpretation. We can easily imagine that it is easier to label a pixel when viewed with its neighboring pixels. This setup is analogous to the improvements a spatial-contextual classifier, like a CNN applying a patch classification, approach bring over a simple pixel-based classifier. But we can also see that it is easier to label a pixel when, aside from viewing its neighboring pixels, its surrounding pixels' labels are given. With contextual label information, the classifier can learn and leverage class spatial co-occurrences. Additionally, we observe that adding more FuseNet instances to the ReuseNet until $\text{\ninstances}=4$ increases the score of all metrics, except for the average class accuracy where ReuseNet-3 marginally outperforms ReuseNet-4. Adding one more instance only improves the F1 score and degrades the other three metrics.
 We can interpret this addition of FuseNet instances as a way to increase ReuseNet's capacity to refine contextual label information fed to it as latter FuseNet instances receive more refined labels.

Figure \ref{fig:reusenet_plots} shows classification results of the best performing ReuseNet, the baseline method FuseNet+CRF, and the plain FuseNet. Both FuseNet+CRF and ReuseNet instances improves the quality of the resulting classified map by producing more regularized classification. We also observe that locations of the errors are carried over from the results of the FuseNet classifier from which both FuseNet+CRF and ReuseNet are based from. However, the occurrences of the errors are diminished especially on the facades of the large buildings. Detection of isolated cars in roads were also improved. Overall, results of ReuseNet-4 show better-quality classified maps by reducing noise in the classification (such as island of impossibly small buildings), further improving delineation of classes with irregular boundaries, and reducing misclassification in regions with ambiguous spectral characteristics such as facades and rooftops of high rise buildings.

\subsubsection{ISPRS Vaihingen dataset}

\begin{table}[!t]
\centering
\caption{Comparison of map regularization approaches on ISPRS Vaihingen dataset}
        \begin{threeparttable}
        \begin{tabular}{lcccc}
		\hline
		\hline
		Network			 		& OA (\%) & $\kappa$ (\%) & AA (\%) & F1 (\%) \\
		\hline
		 No-downsampling FCN \cite{Sherrah2016} & 87.17 & --.-- & --.-- & --.-- \\
		 CNN-FPL \cite{Volpi2017} & 87.83 & 83.83 & 81.35 & 83.58 \\
		 AllConvNet & 86.98 & 82.71 & 87.17 & 85.46 \\
		 FCN in \cite{Sherrah2016} with CRF & 87.90 & --.-- & --.-- & --.-- \\
		 ReuseNet-2 & 87.11 & 82.89 & 85.09 & 85.38 \\
		 ReuseNet-3 & \textbf{88.08} & \textbf{84.18} & \textbf{87.29} & \textbf{87.24} \\
		 ReuseNet-4 & 87.64 & 83.59 & 87.18 & 86.81 \\
		\hline
		\hline
        \end{tabular}
		\begin{tablenotes}
		    \item Unreported values in the reference are denoted by "--.--"
		\end{tablenotes}
        \end{threeparttable}
        \label{tab:regularization_comparison_additional}
\end{table}

\begin{figure*}[tb]
\centering
\includegraphics[width=0.8\textwidth]{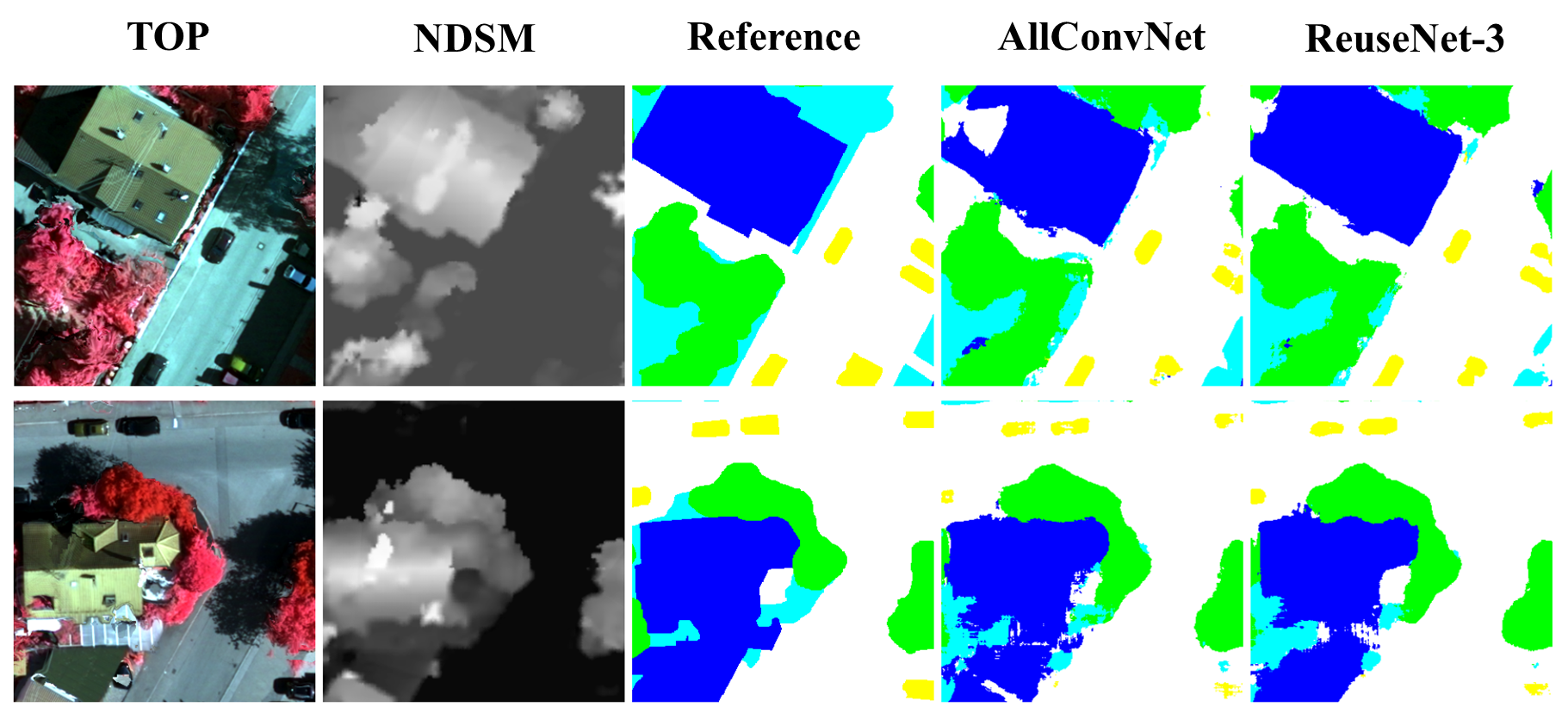}
\caption{Two subsets from the validation tiles of ISPRS Vaihingen dataset showing, from right to left, the true orthophoto, normalized dsm, reference image, and classification maps from AllConvNet and ReuseNet-3 (best performing ReuseNet in this dataset). All ReuseNets reported are ``plain'' meaning initial score maps are filled with zeros. Reusenet-\ninstances\ denotes a ReuseNet composed of \ninstances\ number of FuseNet instances.}
\label{fig:reusenet_plots_additional}
\end{figure*}

Table \ref{tab:regularization_comparison_additional} shows the accuracies obtained by comparing different classification techniques on the ISPRS dataset. These results are in agreement with the results from the previous dataset. All the ReuseNet versions of AllConvNet improve the resuslts on all the four metrics except for AA and F1 of ReuseNet-2 (2.08\% in AA and 0.06\% in F1 respectively). ReuseNet-3, the best performing network, considerably improves all the numerical results of the plain AllConvNet by 1.1\% in OA and is comparable and even greater than the 0.73\% gain after a post-classification CRF in \cite{Sherrah2016}, 1.47\% in $\kappa$, 0.12\% in AA, and 1.78\% in F1. ReuseNet-3 also outperforms best results reported in both \cite{Sherrah2016} and \cite{Volpi2017}.

These results reconfirm that introducing contextual label information through recurrence in an FCN applying a full-patch labeling approach can improve the classification of a VHR image. Similarly, qualitative improvements---such as holes in building being filled, better delineation of all classes in general, lesser artifacts---in the resulting classified maps are observed when ReuseNet is applied as shown in Figure \ref{fig:reusenet_plots_additional}.

\subsubsection{Different initializations}

\begin{figure}[!htb]
\includegraphics[width=0.5\textwidth]{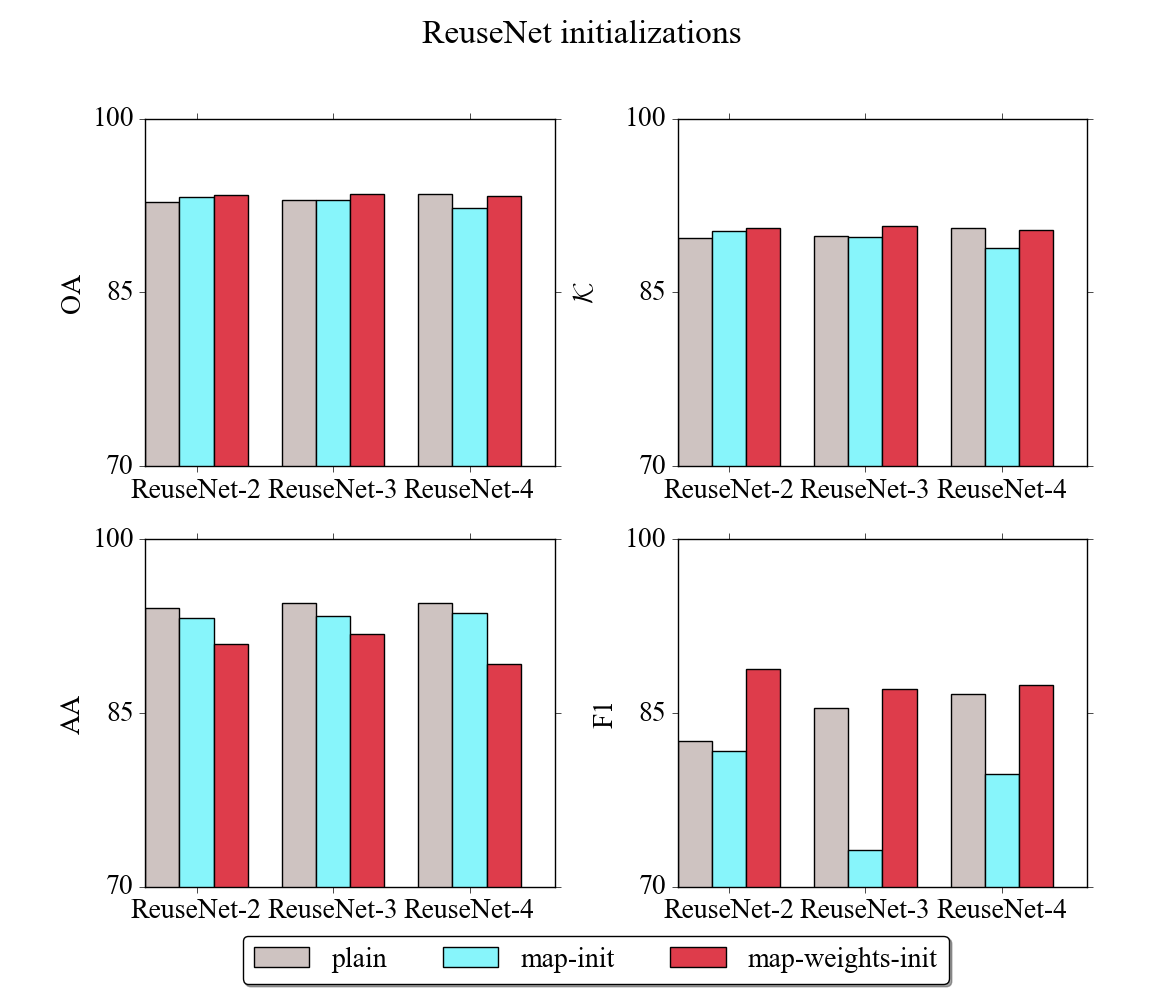}
\caption{Plots showing results of quantitive metrics comparing different ReuseNet initializations; plain, map-init, and map-weights-init correspond to intializing the ReuseNet with zero-score maps, scores from a previously-trained FuseNet, and scores and weights from a previously-trained FuseNet respectively.}
\label{fig:reusenet_inits}
\end{figure}

Figure \ref{fig:reusenet_inits} shows results of quantitive metrics on the three different ReuseNet initializations. There is low variation in the OA and $\kappa$. The trend of the two global scores is also inconsistent across the ReuseNet instances. For ReuseNet-2 and ReuseNet-3, the scores increases marginally (around 0.5\% for OA and 0.8\% for $\kappa$) when initialized with both the scores and weights from a previously-trained FuseNet. But for ReuseNet-4, there is a minor drop in both the scores (around 0.2\% for both scores) when the two intialization methods are introduced. This could mean that increasing the FuseNet instances to a certain amount already provides enough room to a ReuseNet for ``label refinement'' such that gains from the initialization methods are compensated.

Introducing both initialization methods to a ReuseNet degrades the AA by around 0.9--5.2\%. Applying only the initialization using scores from a FuseNet instance (map-init) degrades the F1 by around 0.9--12.2\%. Interestingly, the F1 improve by around 0.8--6.1\% when both initialization methods are introduced (map-weights-init). Decrease in AA can only imply an increase in false positive predictions in most of the classes; while increase in F1 could either mean decrease in false positive predictions or decrease in false negative predictions or both in most of the classes. The results therefore show that the initialization methods promote higher recall rate (decrease in false negatives) in underrepresented classes such as cars.

\subsection{Sensitivity Analysis}

\begin{figure}[tb]
\centering
\includegraphics[width=0.45\textwidth]{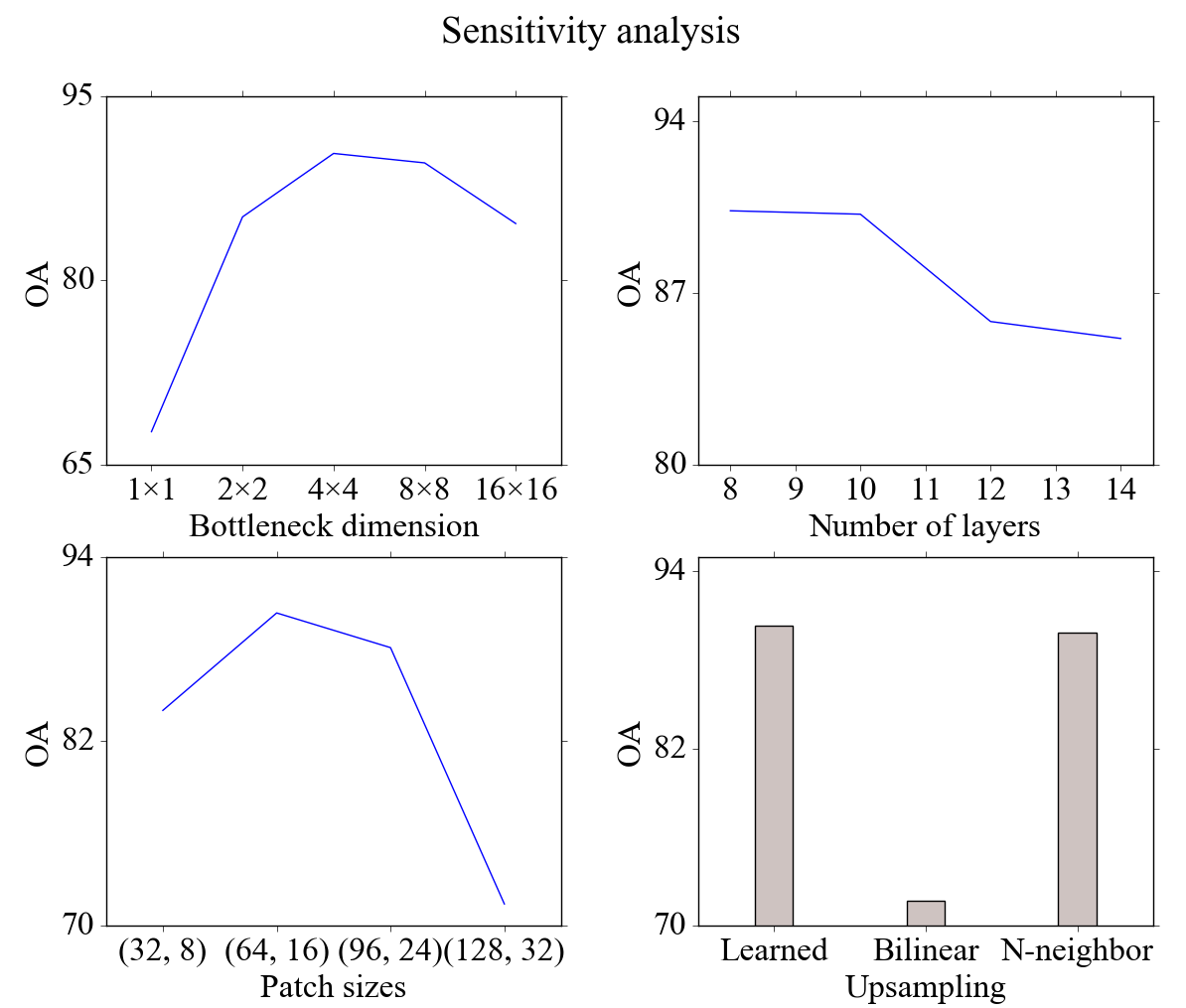}
\caption{Plots showing the results of sensitivity analysis. Patch sizes are written as ``$\langle$(4\patchsize, \patchsize)$\rangle$''. N-neighbor denotes nearest neighbor interpolation.}
\label{fig:sensitivity}
\end{figure}

Fig. \ref{fig:sensitivity} shows the results of the sensitivity analysis performed on four chosen hyperparameters of FuseNet: 1) bottleneck feature map dimensions, 2) number of convolutional layers (in the downsampling part of the network), 3) input patch sizes, and 4) upsampling methods. We got the highest validation accuracy of 90.35\% using a bottleneck feature map dimension of 4\by4 pixels. Decreasing the dimension more than the optimal we found severely degrades the classification resulting to large uniform areas producing stamp-like patterns (especially for 1\by1). Increasing the dimensions produces much noisier classification. Fixing the bottleneck size dimension to 4\by4 and further increasing the number of convolutional layers (without downsampling) did not produce any improvements in the validation accuracy. Increasing the number of these convolutional layers within the bottleneck feature maps effectively increases the receptive field (footprint size in the input layer containing the PAN image patch) of the succeeding units by at least half of the size of kernels used in the convolutional layers. Hence, the results show that: with only eight convolutional layers (with downsampling), we can learn enough contextual information for accurate classification.

We found the optimal patch sizes of 64\by64 for the \inputpan\ and 16\by16 for \inputms. Further increasing the patch sizes results in overclassification of a single class (impervious surface).
Increasing the patch size also increases the proportion of frequently occurring classes in the training sample, possibly resulting into overclassification. Whereas, decreasing the patch size limits the contextual information incorporated in the input, and, hence, can degrade the classification results. Lastly, we find using transposed convolution for learned upsampling to perform better than using interpolation for fixed upsampling (bilinear and nearest neighbor). This result supports the expected flexibility of empirically learning the upsampling operation directly from data.

\section{Conclusion and Future Works}

In this paper, we presented a recurrent multiresolution convolutional network named ReuseNet to classfiy VHR satellite images. The operations for fusing the bands with different resolutions are learned within convolutional layers with corresponding downsampling and upsampling operations to match the resolution of the images. Regularization of the resulting classified maps is achieved by incorporating contextual label information through the recurrent architecture of ReuseNet. Additionally, we investigated various ways to initialize ReuseNet. The effect of varying a set of chosen network hyperparameters to the classification accuracy of the network was explored. Both numerical and qualitative results show the advantages of incorporating image resolution matching and contextual label learning within the training of the classifier. To this end, we provided a single-stage classification pipeline incorporating image fusion, feature extraction, and map regularization, all combined in a convolutional network trained in an end-to-end manner.

We designed the presented network architecture such that it can easily be adapted to other multiresolution image datasets. Inclusion and leverage of contextual label information is also separate from the design of the fusion network in the sense that it can be implemented on network classifying single-resolution images. For future work, we plan to fuse images from different sensors (e.g. Sentinel-2) and classify classes of higher abstraction such as land use instead of land cover.

\ifCLASSOPTIONcaptionsoff
  \newpage
\fi

\bibliographystyle{IEEEtran}

\bibliography{References}

\begin{thebibliography}{10}
\providecommand{\url}[1]{#1}
\csname url@samestyle\endcsname
\providecommand{\newblock}{\relax}
\providecommand{\bibinfo}[2]{#2}
\providecommand{\BIBentrySTDinterwordspacing}{\spaceskip=0pt\relax}
\providecommand{\BIBentryALTinterwordstretchfactor}{4}
\providecommand{\BIBentryALTinterwordspacing}{\spaceskip=\fontdimen2\font plus
\BIBentryALTinterwordstretchfactor\fontdimen3\font minus
  \fontdimen4\font\relax}
\providecommand{\BIBforeignlanguage}[2]{{%
\expandafter\ifx\csname l@#1\endcsname\relax
\typeout{** WARNING: IEEEtran.bst: No hyphenation pattern has been}%
\typeout{** loaded for the language `#1'. Using the pattern for}%
\typeout{** the default language instead.}%
\else
\language=\csname l@#1\endcsname
\fi
#2}}
\providecommand{\BIBdecl}{\relax}
\BIBdecl

\bibitem{Haralick1973}
\BIBentryALTinterwordspacing
R.~Haralick, K.~Shanmugan, and I.~Dinstein, ``{Textural features for image
  classification},'' pp. 610--621, 1973. [Online]. Available:
  \url{http://dceanalysis.bigr.nl/Haralick73-Textural features for image
  classification.pdf}
\BIBentrySTDinterwordspacing

\bibitem{Ojala2002}
T.~Ojala, M.~Pietikainen, and T.~Maenpaa, ``Multiresolution gray-scale and
  rotation invariant texture classification with local binary patterns,''
  \emph{IEEE Transactions on Pattern Analysis and Machine Intelligence},
  vol.~24, no.~7, pp. 971--987, Jul 2002.

\bibitem{DallaMura2010}
M.~D. Mura, J.~A. Benediktsson, B.~Waske, and L.~Bruzzone, ``Morphological
  attribute profiles for the analysis of very high resolution images,''
  \emph{IEEE Transactions on Geoscience and Remote Sensing}, vol.~48, no.~10,
  pp. 3747--3762, Oct 2010.

\bibitem{Fauvel2013}
M.~Fauvel, Y.~Tarabalka, J.~A. Benediktsson, J.~Chanussot, and J.~C. Tilton,
  ``Advances in spectral-spatial classification of hyperspectral images,''
  \emph{Proceedings of the IEEE}, vol. 101, no.~3, pp. 652--675, March 2013.

\bibitem{LeCun1998}
Y.~Lecun, L.~Bottou, Y.~Bengio, and P.~Haffner, ``Gradient-based learning
  applied to document recognition,'' in \emph{Proceedings of the IEEE}, 1998,
  pp. 2278--2324.

\bibitem{Bergado2016}
J.~R. Bergado, C.~Persello, and C.~Gevaert, ``A deep learning approach to the
  classification of sub-decimetre resolution aerial images,'' in
  \emph{Proceedings of the International Geoscience and Remote Sensing
  Symposium (IGARSS) 2016}, vol. 2016-November, 2016, pp. 1516--1519.

\bibitem{Mboga2017}
N.~Mboga, C.~Persello, J.~Bergado, and A.~Stein, ``Detection of informal
  settlements from vhr images using convolutional neural networks,''
  \emph{Remote Sensing}, vol.~9, no.~11, p. 1106, 2017.

\bibitem{Hubel1962}
D.~H. Hubel and T.~N. Wiesel, ``Receptive fields, binocular interaction, and
  functional architecture in the cat's visual cortex,'' \emph{Journal of
  Physiology (London)}, vol. 160, pp. 106--154, 1962.

\bibitem{Ha2013}
\BIBentryALTinterwordspacing
W.~Ha, P.~H. Gowda, and T.~A. Howell, ``A review of potential image fusion
  methods for remote sensing-based irrigation management: part ii,''
  \emph{Irrigation Science}, vol.~31, no.~4, pp. 851--869, Jul 2013. [Online].
  Available: \url{https://doi.org/10.1007/s00271-012-0340-6}
\BIBentrySTDinterwordspacing

\bibitem{Chen2016}
\BIBentryALTinterwordspacing
L.~Chen, G.~Papandreou, I.~Kokkinos, K.~Murphy, and A.~L. Yuille, ``Deeplab:
  Semantic image segmentation with deep convolutional nets, atrous convolution,
  and fully connected crfs,'' \emph{CoRR}, vol. abs/1606.00915, 2016. [Online].
  Available: \url{http://arxiv.org/abs/1606.00915}
\BIBentrySTDinterwordspacing

\bibitem{Paisitkriangkrai2016}
S.~Paisitkriangkrai, J.~Sherrah, P.~Janney, and A.~van~den Hengel, ``Semantic
  labeling of aerial and satellite imagery,'' \emph{IEEE Journal of Selected
  Topics in Applied Earth Observations and Remote Sensing}, vol.~9, no.~7, pp.
  2868--2881, 2016.

\bibitem{Wang2017}
\BIBentryALTinterwordspacing
L.~Wang, X.~Huang, C.~Zheng, and Y.~Zhang, ``A markov random field integrating
  spectral dissimilarity and class co-occurrence dependency for remote sensing
  image classification optimization,'' \emph{ISPRS Journal of Photogrammetry
  and Remote Sensing}, vol. 128, no. Supplement C, pp. 223--239, 2017.
  [Online]. Available:
  \url{http://www.sciencedirect.com/science/article/pii/S0924271616302787}
\BIBentrySTDinterwordspacing

\bibitem{Sherrah2016}
J.~Sherrah, ``Fully convolutional networks for dense semantic labelling of
  high-resolution aerial imagery,'' \emph{arXiv preprint arXiv:1606.02585},
  2016.

\bibitem{Maggiori2017}
E.~Maggiori, Y.~Tarabalka, G.~Charpiat, and P.~Alliez, ``Convolutional neural
  networks for large-scale remote-sensing image classification,'' \emph{IEEE
  Transactions on Geoscience and Remote Sensing}, vol.~55, no.~2, pp. 645--657,
  Feb 2017.

\bibitem{Persello2017}
C.~Persello and A.~Stein, ``Deep fully convolutional networks for the detection
  of informal settlements in vhr images,'' \emph{IEEE Geoscience and Remote
  Sensing Letters}, vol.~PP, no.~99, pp. 1--5, 2017.

\bibitem{Volpi2017}
M.~Volpi and D.~Tuia, ``Dense semantic labeling of subdecimeter resolution
  images with convolutional neural networks,'' \emph{IEEE Transactions on
  Geoscience and Remote Sensing}, vol.~55, no.~2, pp. 881--893, 2017.

\bibitem{Zhao2017}
W.~Zhao, L.~Jiao, W.~Ma, J.~Zhao, J.~Zhao, H.~Liu, X.~Cao, and S.~Yang,
  ``Superpixel-based multiple local cnn for panchromatic and multispectral
  image classification,'' \emph{IEEE Transactions on Geoscience and Remote
  Sensing}, vol.~55, no.~7, pp. 4141--4156, 2017.

\bibitem{Xu2017}
X.~Liu, L.~Jiao, J.~Zhao, J.~Zhao, D.~Zhang, F.~Liu, S.~Yang, and X.~Tang,
  ``Deep multiple instance learning-based spatial-spectral classification for
  pan and ms imagery,'' \emph{IEEE Transactions on Geoscience and Remote
  Sensing}, vol.~PP, no.~99, pp. 1--13, 2017.

\bibitem{Jordan1997}
M.~I. Jordan, ``Chapter 25 - serial order: A parallel distributed processing
  approach,'' in \emph{Neural-Network Models of Cognition Biobehavioral
  Foundations}, ser. Advances in Psychology, J.~W. Donahoe and V.~P. Dorsel,
  Eds.\hskip 1em plus 0.5em minus 0.4em\relax North-Holland, 1997, vol. 121,
  pp. 471--495.

\bibitem{Goodfellow2016}
I.~Goodfellow, Y.~Bengio, and A.~Courville, \emph{Deep Learning}.\hskip 1em
  plus 0.5em minus 0.4em\relax MIT Press, 2016,
  \url{http://www.deeplearningbook.org}.

\bibitem{Pinheiro2014}
P.~H.~O. Pinheiro and R.~Collobert, ``Recurrent convolutional neural networks
  for scene labeling,'' in \emph{Proceedings of the 31st International
  Conference on Machine Learning (ICML)}, 2014.

\bibitem{Maas2013}
A.~L. Maas, A.~Y. Hannun, and A.~Y. Ng, ``Rectifier nonlinearities improve
  neural network acoustic models,'' in \emph{ICML 2013 Workshop on Deep
  Learning for Audio, Speech, and Language Processing}, 2013.

\bibitem{He2015b}
K.~He, X.~Zhang, S.~Ren, and J.~Sun, ``Delving deep into rectifiers: Surpassing
  human-level performance on {I}magenet classification,'' in \emph{IEEE
  International Conference on Computer Vision (ICCV) 2015}, 2015.

\bibitem{Clevert2016}
D.~Clevert, T.~Unterthiner, and S.~Hochreiter, ``Fast and accurate deep network
  learning by exponential linear units ({ELUs}),'' in \emph{International
  Conference on Learning Representations}, 2016.

\bibitem{Long2015}
J.~Long, E.~Shelhamer, and T.~Darrell, ``Fully convolutional networks for
  semantic segmentation,'' in \emph{{IEEE} Conference on Computer Vision and
  Pattern Recognition, {CVPR} 2015, Boston, MA, USA, June 7-12, 2015}, 2015,
  pp. 3431--3440.

\bibitem{Badrinarayanan2017}
V.~Badrinarayanan, A.~Kendall, and R.~Cipolla, ``Segnet: A deep convolutional
  encoder-decoder architecture for image segmentation,'' \emph{IEEE
  Transactions on Pattern Analysis and Machine Intelligence}, 2017.

\bibitem{Bottou2012}
L.~Bottou, ``Stochastic gradient descent tricks,'' in \emph{Neural Networks:
  Tricks of the Trade}.\hskip 1em plus 0.5em minus 0.4em\relax Springer Berlin
  Heidelberg, 2012, pp. 421--436.

\bibitem{Noh2015}
H.~Noh, S.~Hong, and B.~Han, ``Learning deconvolution network for semantic
  segmentation,'' in \emph{Proceedings of the IEEE International Conference on
  Computer Vision}, 2015, pp. 1520--1528.

\bibitem{Simonyan2014}
K.~Simonyan and A.~Zisserman, ``Very deep convolutional networks for
  large-scale image recognition,'' \emph{arXiv preprint arXiv:1409.1556}, 2014.

\bibitem{Ioffe2015}
S.~Ioffe and C.~Szegedy, ``Batch normalization: Accelerating deep network
  training by reducing internal covariate shift,'' \emph{arXiv preprint
  arXiv:1502.03167}, 2015.

\bibitem{Krahenbuhl2011}
P.~Kr\"{a}henb\"{u}hl and V.~Koltun, ``Efficient inference in fully connected
  crfs with gaussian edge potentials,'' in \emph{Advances in Neural Information
  Processing Systems 24}, J.~Shawe-Taylor, R.~S. Zemel, P.~L. Bartlett,
  F.~Pereira, and K.~Q. Weinberger, Eds.\hskip 1em plus 0.5em minus 0.4em\relax
  Curran Associates, Inc., 2011, pp. 109--117.

\bibitem{Hester2008}
D.~Hester, H.~Cakir, S.~Nelson, and S.~Khorram, ``Per-pixel classification of
  high spatial resolution satellite imagery for urban land-cover mapping,''
  \emph{Photogrammetric Engineering and Remote Sensing}, vol.~74, no.~4, pp.
  463--471, 2008.

\bibitem{Bakir2007}
G.~H. Bakir, T.~Hofmann, B.~Sch\"{o}lkopf, A.~J. Smola, B.~Taskar, and S.~V.~N.
  Vishwanathan, \emph{Predicting Structured Data (Neural Information
  Processing)}.\hskip 1em plus 0.5em minus 0.4em\relax The MIT Press, 2007.

\bibitem{Lafferty2001}
J.~D. Lafferty, A.~McCallum, and F.~C.~N. Pereira, ``Conditional random fields:
  Probabilistic models for segmenting and labeling sequence data,'' in
  \emph{Proceedings of the Eighteenth International Conference on Machine
  Learning}, ser. ICML '01.\hskip 1em plus 0.5em minus 0.4em\relax San
  Francisco, CA, USA: Morgan Kaufmann Publishers Inc., 2001, pp. 282--289.

\bibitem{Giorgi2014}
A.~D. Giorgi, G.~Moser, and S.~B. Serpico, ``Contextual remote-sensing image
  classification through support vector machines, markov random fields and
  graph cuts,'' in \emph{2014 IEEE Geoscience and Remote Sensing Symposium},
  July 2014, pp. 3722--3725.

\bibitem{Cramer2010}
M.~Cramer, ``{The DGPF-test on digital airborne camera evaluation --- overview
  and test design},'' \emph{Photogrammetrie - Fernerkundung - Geoinformation},
  vol.~2, pp. 73--82, 2010.

\bibitem{Mousa2017}
Y.-K. Mousa, P.~Helmholz, and D.~Belton, ``New dtm extraction approach from
  airborne images derived dsm,'' vol.~42, no. 1W1, 2017, pp. 75--82.

\bibitem{Laben2000}
\BIBentryALTinterwordspacing
C.~Laben and B.~Brower, ``Process for enhancing the spatial resolution of
  multispectral imagery using pan-sharpening,'' Jan.~4 2000, uS Patent
  6,011,875. [Online]. Available:
  \url{https://www.google.com/patents/US6011875}
\BIBentrySTDinterwordspacing

\bibitem{He2016}
K.~He, X.~Zhang, S.~Ren, and J.~Sun, ``Deep residual learning for image
  recognition,'' in \emph{Computer Vision and Pattern Recognition (CVPR), 2016
  IEEE Conference on}, 2016.

\bibitem{Glorot2010}
X.~Glorot and Y.~Bengio, ``Understanding the difficulty of training deep
  feedforward neural networks.'' in \emph{Aistats}, vol.~9, 2010, pp. 249--256.

\bibitem{Zeiler2012}
M.~D. Zeiler, ``Adadelta: An adaptive learning rate method,'' \emph{CoRR}, vol.
  abs/1212.5701, 2012.

\end{thebibliography}

\begin{IEEEbiography}{John Ray Bergado}
Biography text here.
\end{IEEEbiography}

\begin{IEEEbiographynophoto}{Claudio Persello}
Biography text here.
\end{IEEEbiographynophoto}

\begin{IEEEbiographynophoto}{Alfred Stein}
Biography text here.
\end{IEEEbiographynophoto}

\end{document}